\documentclass[journal,twoside]{IEEEtran}
\ifCLASSINFOpdf
\else
\fi

\usepackage{amsmath}
\DeclareMathOperator*{\argmax}{arg\,max}

\usepackage{bm}
\usepackage{amssymb}
\newcommand{\X}{\mathbf{X}}
\newcommand{\Y}{\mathbf{Y}}
\newcommand{\x}{\mathbf{x}}
\newcommand{\y}{\mathbf{y}}
\usepackage[ruled,linesnumbered]{algorithm2e}

\usepackage{array}
\usepackage{url}

\ifCLASSOPTIONcompsoc
  \usepackage[caption=false,font=normalsize,labelfont=sf,textfont=sf]{subfig}
\else
  \usepackage[caption=false,font=footnotesize]{subfig}
\fi
\usepackage{rotating}

\usepackage[pagebackref=true,breaklinks=true,letterpaper=true,colorlinks,bookmarks=false]{hyperref}
\usepackage{multirow}
\usepackage{booktabs}
\usepackage{comment}
\usepackage{graphicx}
\usepackage{graphics}

\begin{document}
%
\title{Solving Inverse Computational Imaging Problems \\ using Deep Pixel-level Prior}
%
%
%

\author{Akshat~Dave,~Anil~Kumar~Vadathya,~Ramana~Subramanyam,~Rahul~Baburajan,~Kaushik~Mitra,~\IEEEmembership{Member, ~IEEE} 
\thanks{Akshat Dave is with the Department of Electrical and Computer Engineering, Rice University, Houston, TX, USA (e-mail: akshat.dave@rice.edu).}
\thanks{Anil Kumar Vadathya, Rahul Baburajan and Kaushik Mitra are with the Computational Imaging Lab, Indian Institute of Technology (IIT) Madras, Chennai, India.}
\thanks{Ramana Subramanyam is with Insight Centre for Data Analytics, Dublin, Ireland.}}

%
%

\markboth{Journal of \LaTeX\ Class Files,~Vol.~XX,~No.~XX,~April~20XX}{Dave \MakeLowercase{\textit{et al.}}: Solving Inverse Computational Imaging Problems using Deep Pixel-level Prior}
%



\maketitle


\begin{abstract}
Signal reconstruction is a challenging aspect of computational imaging as it often involves solving ill-posed inverse problems. Recently, deep feed-forward neural networks have led to state-of-the-art results in solving various inverse imaging problems. However, being task specific, these networks have to be learned for each inverse problem. 
On the other hand, a more flexible approach would be to learn a deep generative model once and then use it as a signal prior for solving various inverse problems. 
We show that among the various state of the art deep generative models, autoregressive models are especially suitable for  our purpose for the following reasons. First, they explicitly model the pixel level dependencies and hence are capable of reconstructing low-level details such as texture patterns and edges better. Second, they provide an explicit expression for the image prior which can then be used for MAP based inference along with the forward model. Third, they can model long range dependencies in images which make them ideal for handling global multiplexing as encountered in various compressive imaging systems. 
We demonstrate the efficacy of our proposed approach in solving three computational imaging problems: Single Pixel Camera (SPC), LiSens and FlatCam. For both real and simulated cases, we obtain better reconstructions than the state-of-the-art methods in terms of perceptual and quantitative metrics.

\end{abstract}


%
\IEEEpeerreviewmaketitle

\begin{figure*}
    \centering
    \includegraphics[width=1\textwidth]{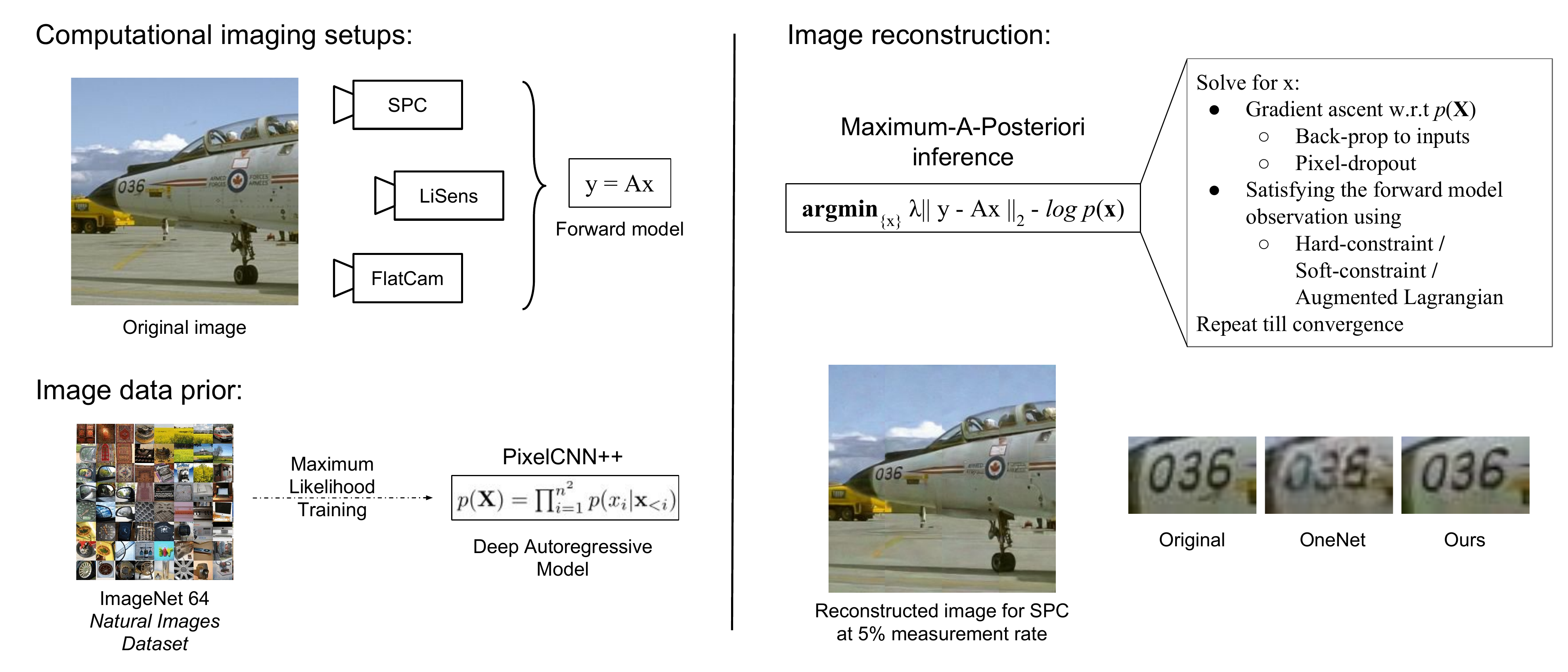}
    \caption{An overview of our approach. We employ a single deep autoregressive model learned on natural images for solving multiple inverse problems. From the zoomed in patch of the reconstructed image in the inset it is evident that our approach has better pixel-level consistencies as compared to existing latent representation based models like OneNet \cite{chang2017one}.}
    \label{fig:main_fig}
\end{figure*}

\section{Introduction}

\IEEEPARstart{C}{omputational} imaging systems enable us to extract much more information out of the visual world as compared to the traditional imaging systems. This is achieved by jointly designing optics, to encode the desired signal information, and algorithms to reconstruct the signal back from those measurements. Signal reconstruction corresponds to inverting the forward model used in acquiring the measurements. Hence, reconstruction algorithms for different computational imaging devices amount to solving different inverse problems. Solving these inverse problems becomes challenging as they are often ill-posed. For  compressive imaging setups such as Single Pixel Camera (SPC) \cite{duarte2008single}, \cite{wang2015lisens}, high speed imaging \cite{reddy2011p2c2},  \cite{hitomi2011video} and compressive hyper-spectral imaging \cite{wagadarikar2008single}, the reconstruction becomes ill-posed as the number of measurements is quite less than the signal dimension. 

Generally, for solving an ill-posed problems, we need to incorporate the prior information about the signal to be reconstructed. Traditionally these priors are either analytically derived or hand-crafted based on the observations. For example, sparsity of image gradients \cite{rudin1992nonlinear}, sparsity of coefficients in wavelet and DCT domain \cite{portilla2003image} etc. have been used for solving inverse imaging problems. However, the underlying data distribution may not precisely follow these analytic priors leading to poor solutions in challenging scenarios. Dictionary learning \cite{aharon2006rm} methods being data driven are an improvement over these analytic priors. However, being limited by patch size they cannot account for long range dependencies which are necessary for handling global multiplexing in case of compressive image reconstruction.



On the other hand, deep learning based reconstruction algorithms recently have led to state-of-the-art results in solving such ill-posed problems in computational imaging \cite{kulkarni2016reconnet}, \cite{yao2017dr} \cite{metzler2017learned} \cite{sinha2017lensless}. These approaches typically learn an inverse mapping from measurements to the signal by minimizing reconstruction loss on a set of training examples. However, this kind of training, popularly known as discriminative learning, makes the network task specific. 
Furthermore, we need to retrain the network for various parameter settings of the forward model. For example, for every new setting of measurement rate and sensing matrix in SPC, we need to relearn the network parameters.  
Instead of having to design/retrain a different network for each task and parameter setting, it would be more efficient to have a generalized framework which can be used for solving various inverse problems.

A more flexible approach would be to learn the natural image statistics using a generative model and use it for solving various inverse problems. 
Recently, deep generative models especially using \textit{autoregressive framework} \cite{van2016pixel, theis2015generative,salimans2017pixelcnn++} have led to state-of-the-art performance in modeling natural image manifold. Autoregressive models factorize the image distribution as a 2D directed causal graph and hence model it as a $2$-D sequence where current pixel's distribution is conditioned on the causal context. 
By employing deep neural networks for summarizing the causal context, autoregressive models excel at capturing long range dependencies in images. Also, being a pixel level model it explicitly accounts for higher order correlations like texture patterns, sharp edges, etc. within a neighbourhood. Thus, these models are capable of generating visually convincing and crisp images \cite{van2016pixel}. Examples of deep autoregressive image models are recurrent image density estimator (RIDE) \cite{theis2015generative}, pixel recurrent neural networks (PixelRNN) and its CNN equivalent (PixelCNN) \cite{van2016pixel} and PixelCNN++ \cite{salimans2017pixelcnn++}. 

We show that deep autoregressive generative models are ideally suitable for solving various computational imaging problems for the following reasons. First, it explicitly models the distribution of each pixel in relation to its causal neighbor. Thus, when used as an image prior, this explicit pixel dependency modeling helps it to better reconstruct low level details without artifacts (see Figure \ref{fig:main_fig}). Second, this framework gives us an explicit expression for the image prior, which can be used for doing MAP inference. Moreover, the entire framework is differentiable, which is amenable for gradient based inference. Third, its ability to capture long range dependencies in images makes them ideal for handling global multiplexing in compressive imaging setups. 
Given these advantages with deep autoregressive models, we use it for solving various computational imaging problems such as - Single Pixel Camera (SPC) \cite{duarte2008single}, Line Sensor (LiSens) \cite{wang2015lisens} and lensless imaging - FlatCam \cite{asif2017flatcam}. Our results demonstrate that we perform better than the current state-of-the-art methods in both traditional and learning based approaches. 

In summary we make the following contributions: 
\begin{itemize}
    \item We propose a versatile approach which employs the same learned prior model for solving various computational imaging problems. 
    
    \item We propose to use a deep autoregressive model, PixelCNN++, as an image prior. The autoregressive nature of this prior ensures pixel-level consistencies in the reconstruction and hence provides better quality than using latent representation based models such as OneNet \cite{chang2017one} as shown in Figure \ref{fig:main_fig}.
    
    \item We utilize back-propagation to the inputs for obtaining tractable estimates of the prior gradients and employ them for solving inverse problems using MAP inference.
    
    \item We observe that randomly dropping the gradient updates for a certain percentage of pixels at every iteration helps in reconstructing the texture better. We analyze the effect of this pixel dropout ratio on the quality of reconstructions.
    
    \item We demonstrate better reconstructions than the existing state-of-the-art methods for three computational imaging problems: Single Pixel Camera, LiSens, and FlatCam. 
\end{itemize}

\section{Related Work}

\textbf{Compressive imaging} Single Pixel Camera (SPC) \cite{duarte2008single} is a classic example of compressive imaging. It uses a programmable digital micro-mirror device (DMD) array to multiplex the scene on to a single photodetector. Using different settings on the DMD, we can sequentially acquire a set of measurements. Thus, scene at full resolution is reconstructed from much less than 100\% measurements. Compressive imaging systems pose a viable solution for high resolution imaging in non-visible parts of the spectrum where full frame sensors are very expensive. 

The measurement bandwidth of the SPC is limited by the operating speed of the DMDs (Tens of kHz for commercially-available units). With this speed, SPC cannot be extended for high resolution video sensing. On one end, we have exorbitant full frame sensors (Nyquist sampling) for high resolution imaging in non visible bands, and on the other, we have SPC, an inexpensive compressive sensing setup but with low measurement rates. Wang et al. \cite{wang2015lisens} propose LiSens - Line Sensor based compressive camera which lies midway between these two imaging extremes. Each pixel in the line sensor is mapped to a row in DMD array. Thus, unlike SPC, where the whole scene is multiplexed, here only rows of the scene are multiplexed. 

\textbf{Lensless imaging} FlatCam \cite{asif2017flatcam} and DiffuserCam \cite{antipa20173d} are novel imaging systems which get rid of the conventional lens optics. Instead, they use amplitude and diffuser mask respectively to encode light coming from different parts of the scene onto the sensor. As a result, information localized at a point in the scene gets spread throughout the sensor, making priors essential for accurate recovery of the image. These works use traditional reconstruction algorithms such as Total Variation norm and Tikhonov regularization which are quick but do not provide natural looking reconstructions. 

\textbf{Reconstruction with analytical priors} Many algorithms have been proposed for compressive image reconstruction. Typically, reconstruction algorithms use $l_1$ regularization, exploiting the sparsity of spatial gradients in natural images. Total Variation (TV) minimization prior \cite{rudin1992nonlinear, chambolle2004algorithm} is the most commonly used reconstruction algorithm based on this sparsity. Chengbo et al. \cite{li2013efficient} propose an efficient augmented Lagrangian based TV minimization for CS reconstruction. Recent approaches involving compressive architectures such as fpa-cs \cite{chen2015fpa}, LiSens \cite{wang2015lisens}, and video CS \cite{sankaranarayanan2012cs}, demonstrated successful results with TV minimization prior. However, at lower measurement rates, reconstructions suffer from the piece-wise smooth modeling of TV prior and results tend to be blocky, as is noted by recent works \cite{kulkarni2016reconnet, dave2017compressive}. Metzler et al. \cite{metzler2016denoising} propose a denoiser based CS reconstruction algorithm. Specifically, use a Gaussian denoiser with approximate message passing algorithm (D-AMP). At very low measurement rates, the denoiser tends to result in overly smooth images as is recently shown by Dave et al. \cite{dave2017compressive}, Kulkarni et al. \cite{kulkarni2016reconnet}.

\textbf{Data driven CS reconstruction} Duarte et al. \cite{duarte2009learning} propose an approach for simultaneous learning of the sensing matrix and dictionary atoms. Due to the small patch size of the atoms, their usage for compressive image reconstruction is limited to local multiplexing, unlike the actual SPC involving global multiplexing of the scene. Reconstruction algorithms using convolutional neural networks (CNNs) typical take input as measurements from an image patch and try to output the image back by minimizing the reconstruction loss. Kulkarni et al. \cite{kulkarni2016reconnet} proposed ReconNet, Yao et al. \cite{yao2017dr} proposed $DR^2$-Net having residual connections for reconstruction. Although these approaches lead to a non-iterative and hence faster inference, being task specific, they only work for the fixed settings of the sensing matrix and measurement rates used for training. Changing the settings requires retraining the architecture which is not very appealing. Also, being patch-wise, they also fail to account for global multiplexing in SPC. 

\textbf{Deep generative models}
With the success of deep neural networks, there have been multiple works proposing deep generative models, which explicitly or implicitly try to model the distribution of natural images. For example, latent representation models like adversarial networks, GAN by Goodfellow et al. \cite{goodfellow2014generative}, variational auto-encoders by Kingma et al. \cite{kingma2013auto} and autoregressive models like RIDE by Theis et al. \cite{theis2015generative}, PixelRNN/CNN by Oord et al. \cite{van2016pixel}, PixelCNN++ by Salimans et al. \cite{salimans2017pixelcnn++}. GANs learn to transform samples from a Gaussian distribution to a sample in the natural image manifold via a generator network, which is trained with an adversarial learning framework involving a discriminator network. VAEs are a probabilistic framework of autoencoders that learn to encode and decode the images from a distribution. 

Autoregressive models factorize an image as a 2D directed graph by conditioning the current pixel $x_i$'s distribution on the pixels before it as in a raster scan $x_{<i}$. Modeling this conditional density is analogous to sequence modeling and initial methods proposed to use spatial 2D recurrent neural networks, given their efficacy in modeling sequences. RIDE by Theis et al. \cite{theis2015generative} uses 2D Long Short Term Memory (LSTM) units called Spatial-LSTMs for modeling the causal context $x_{<i}$, and GSMs for parametrizing the distribution. PixelRNN by Oord et al. \cite{van2016pixel} uses a much complex architecture using LSTMs and residual connections to better handle the causal context. Importantly, it models the conditional density as a discrete distribution with $x_i \in \{0, 1, \ldots 255\}$. PixelRNN has resulted in state-of-the-art negative loglikelihood (NLL) scores. However, due to the sequential nature of distribution modeling, both training and sampling are computationally demanding with the runtime as $O(N)$, where $N$ is the total number of pixels. Oord et al. proposed PixelCNN which is a convolutional version of PixelRNN. This led to an improvement in
the training time by a large factor at the cost of slight loss in the accuracy as with convolutions we can now only capture bounded context. Salimans et al. \cite{salimans2017pixelcnn++} proposed PixelCNN++, which builds on PixelCNN by employing a discretized mixture of logistics for modeling the distribution, and using drop-out regularization, and additional skip connections. It improves on the NLL score over PixelRNN on the CIFAR dataset leading to state-of-the-art results. 


\textbf{Deep image priors} When solving linear inverse problems using the alternating direction method of multipliers (ADMM) algorithm, Venkatakrishnan et al. \cite{venkatakrishnan2013plug} observed that it results in two decoupled optimizations. The first one enforces the data prior while the second enforces data fidelity to the observation. The first step can be thought of as a denoising problem, thus, a denoiser can be employed to solve this step thereby avoiding the need for an explicit image prior. Venkatakrishnan et al. \cite{venkatakrishnan2013plug} use denoisers like BM3D \cite{dabov2009bm3d} in ADMM setting for image restoration. Inspired by this, recent methods propose learning-based proximal operators for the denoising step of ADMM. OneNet by Chang et al. \cite{chang2017one}, CNN denoiser by Zhang et al.\cite{zhang2017learning}, Meinhardt et al.  \cite{meinhardt2017learning} . In this work, we compare our explicit natural image prior based MAP inference with the learned proximal operator of OneNet. Our evaluations show that our results are superior to OneNet. It is important to note that OneNet's proximal operator uses adversarial loss \cite{goodfellow2014generative} which is known to result in sharper recovery of details.

In this paper, we extend upon our previous work, Dave et al. \cite{dave2017compressive} (RIDE-CS), where we used recurrent image density estimator (RIDE) for CS reconstruction. We observed that the sequential nature of recurrent networks in RIDE makes it too slow for inference and training (computational cost is proportional to the image size). Also, in our experiments, the two layer RIDE fails to yield results comparable to recent approaches like OneNet \cite{chang2017one}. Here, we explore sophisticated deep autoregressive models which are order faster than RIDE-CS for both training and inference. We apply the deep autoregressive model based inference to recent frameworks in computational imaging like LiSens \cite{wang2015lisens} and FlatCam \cite{asif2017flatcam}. We enhance the inference algorithm by incorporating the augmented Lagrangian method when necessary. 
In addition, we improve texture recovery using pixel-wise stochastic gradient updates.

\section{Inference with deep autoregressive models}

\subsection{Problem Formulation}
Consider $\X$ to be a $n\times n$ matrix corresponding to a natural image and $f$ to be a linear transformation corresponding to the forward model of a computational camera. The measurements obtained $\Y$ can be written as $\Y = f(\X)$. Our goal is to reconstruct back the image $\X$ from the measurements $\Y$. 

Discriminative networks learn the inverse mapping $\hat{\X} = g(\Y)$ by modelling $g$ as a deep neural network and minimizing the reconstruction error on a set of training examples $\{\X_i,\Y_i\}$. Hence, the inverse mapping is implicitly dependant on the forward model $f$. Dealing with reconstructions for multiple forward models would require learning separate networks for each model which can be expensive. 

For our generative approach, we model the distribution of natural images $p(\X)$ using a deep autoregressive model.  
We formulate the inverse problem as MAP inference. Hence, the estimated image $\mathbf{\hat{X}}$ can be written as 
\begin{align}
\mathbf{\hat{X}} &= \argmax_{\X} log(p(\X|\Y))\\
&= \argmax_{\X} ( log(p(\Y|\X)) + log(p(\X)))
\end{align}
The likelihood term $p(\Y|\X)$ varies for different imaging systems based on the forward model but the image prior $p(\mathbf{\X})$ remains the same. Thus, we need to learn the prior only once for all the problems.

\subsection{Forward Models}
Let the $n^2 \times 1$ column vector $\mathbf{x}$ represent the rasterized version of the $n \times n$ image matrix $\X$ i.e. $\mathbf{x} \triangleq vec(\X)$ by taking pixels row by row. The forward models that we consider in this work are as follows:
\subsubsection{Randomly Missing Pixels}
Here, we randomly set certain number of pixels in an image to by missing, by setting their values to zero. Hence, $\y$ i.e. the vectorized version of the resultant image can be written as 
\begin{equation}
\y = \mathbf{m} \circ \x
\end{equation}
where $\circ$ denotes the Hadamard product and $\mathbf{m}$ is a Bernoulli random vector. The above equation can also be expressed in a matrix-vector multiplication form as :  
\begin{equation}
\y = \mathbf{M} \x
\end{equation}
where $\mathbf{M}$ is a sub-sampling matrix. 

\subsubsection{Single Pixel Camera}
In SPC \cite{duarte2008single}, the DMD array optically multiplexes the scene onto a single pixel sensor. By changing the orientation of the array, we will get different multiplexing patterns, which results in different measurements. If $\y$ is the vector of $m$ single pixel measurements from SPC and $\Phi$ is the $m \times n^2$ compressive sensing matrix, then we have the forward model as:
\begin{equation}
\y = \bm{\Phi} \x.
\end{equation}

\subsubsection{LiSens}
In Lisens \cite{wang2015lisens}, the 2D image of the scene formed on the DMD plane is mapped onto a 1D line-sensor which essentially captures the 1D integral of the 2D image (along rows or columns). If $Y$ is the $m \times n$ matrix formed by stacking $m$ line sensor measurements from Lisens and $\Phi$ is the $m \times n$ sensing matrix, then we have
\begin{equation}
\Y = \bm{\Phi} \X
\end{equation}

\subsubsection{FlatCam}
FlatCam \cite{asif2017flatcam} replaces the lens system by a coded amplitude mask close to the sensor. For ease of calibration, this mask is designed to be separable, i.e., it can be written as an outer product of 2 one dimensional patterns. Neglecting the diffraction effects, it was shown in \cite{asif2017flatcam} that using such a mask, the $m \times m$ measurements $\Y$ obtained on the FlatCam sensor can be written as 
\begin{equation}
\Y = \bm{\Phi_L} \X \bm{\Phi_R}^T
\end{equation}
where $\bm{\Phi_L}$ and $\bm{\Phi_R}$ are $m \times n$ matrices corresponding to 1-D convolution of the scene $\X$ along the rows and columns respectively.

\subsection{Deep autoregressive model}
Here we model the dependencies between pixels using a directed probabilistic chain. The pixel $x_i$ depends on all the pixels before the index $i$ in $\x$, which we denote as $\mathbf{x}_{<i}$. Hence the joint distribution over the pixels in the image can be factorized as 

\begin{equation}
    p(\X) = p(x_1,x_2,\dots,x_{n^2}) = \prod_{i = 1}^{n^2} p(x_i|\x_{<i})
\end{equation}
In this work, we use state-of-the-art autoregressive generative model, PixelCNN++ \cite{salimans2017pixelcnn++}. Here, the context $\mathbf{x}_{<i}$ for the conditional distribution of each of the pixels is modelled using a deep convolutional neural network with residual connections. The convolution kernels are masked appropriately to ensure that the context of a pixel does not depend on the pixels after it. The conditional distribution is then modelled as a mixture of logistic distributions, where the parameters of the distribution depend on the context. This model is then learned on RGB images using maximum likelihood training.

Once the model is trained, it can be used to solve different inference tasks, as we describe below. Sampling from autoregressive models is slow because of their sequential nature which limits their utility. However, for our approach, we only require the gradients of the density $p(\X)$ with respect to the image $X$. This can be computed efficiently using backpropagation to the inputs.

\section{Optimization methods for deep autoregressive inference}
In this section, we discuss inference methods for various forward models discussed earlier. We want the desired solution to have higher likelihood (lower NLL) under the image prior and at the same time satisfy the constraints specified by the forward model. For this, we perform projected gradient descent. We divide our approach into three categories based on the amount of noise and the kind of forward model. Hard constraint (equality) method is used when there is less or no measurement noise (Section \ref{sec:hard_con}). For certain imaging models like FlatCam, there is no closed form for the projection operator. We instead use the Augmented Lagrangian Method (ALM), see Section \ref{sec:augmented_lag}. For the cases of high noise, the measurements deviate significantly from the forward model, and the soft constraint method (inequality) is used (Section \ref{sec:soft_con}). Further, in Sections \ref{sec:stoch_grad} and \ref{sec:splitting}, we describe two implementation hacks which have proved useful for our approach. 

\subsection{Hard constraint method}
\label{sec:hard_con}
We first analyze the case when the measurement is directly obtained using the imaging model without any noise. $\Y$ is then a deterministic function of $\X$ and hence the likelihood term would correspond to constraints. The problem can be formulated as 
\begin{align}
\mathbf{\hat{X}} &= \argmax_{\X} ( log(p(\X)) \text{ such that } \Y = f(\X)
\end{align}
where $f$ is provided by the imaging model. The signal prior model is the learned autoregressive model with parameters $\theta$. Also, we constrain the intensity of the image to be between $0$ and $1$. Thus our problem is given by:
\begin{align}
\mathbf{\hat{X}} &= \argmax_{\X} ( log(p_{\theta}(\X)) \text{ s.t. } \Y = f(\X), 0 \leq \X_{ij} \leq 1
\end{align}
Let $\mathcal{C}_1$ and $\mathcal{C}_2$ denote the constraint sets $\{\X:\Y = f(\X)\}$ and $\{\X: 0 \leq \X_{ij} \leq 1 \quad \forall i,j \}$ respectively.


We use projected gradient descent to solve this constrained optimization, which involves performing the following steps iteratively:
\begin{align}
\label{eq:ascent}
\mathbf{H}_k &= \X_{k} + \alpha \nabla_{\X} log(p_{\theta}(\X_k))\\
\label{eq:project}
\mathbf{J}_k &= \Pi_{\mathcal{C}_1}(\mathbf{H}_k)\\
\label{eq:clip}
\mathbf{X}_{k+1} &= \Pi_{\mathcal{C}_2}(\mathbf{J}_{k})
\end{align}
where $\Pi_{\mathcal{C}_1}$ and $\Pi_{\mathcal{C}_2}$ are projection operators to the constraint sets $\mathcal{C}_1$ and $\mathcal{C}_2$ respectively. For Eq.\ \ref{eq:ascent} backpropagation to the inputs is used to get the data gradients. For Eq.\ \ref{eq:clip}, pixels in the image are clipped between $0$ and $1$ in every iteration. 

$\Pi_{\mathcal{C}_1}$ is different for different imaging problems. For the randomly missing pixels case, 
\begin{align}
\label{eq:proj_inpaint}
\mathbf{j_k} = (\mathbf{1} - \mathbf{m}) \circ \mathbf{h_k} + (\mathbf{m}) \circ \mathbf{y}
\end{align}
where $\mathbf{1}$ is an $n^2$ vector of ones. This implies that we should only be updating the missing pixels and leave the other pixels the same, which is intuitive. 

For Single Pixel Camera we have,
\begin{align}
\label{eq:proj_spc}
\mathbf{j_k} = \mathbf{h}_k-\Phi^T\left(\Phi\Phi^T\right)^{-1}\left(\Phi\mathbf{h}_k-\mathbf{y}\right)
\end{align}
where $\mathbf{j_k}$ and $\mathbf{h}_k$ are vector representations of matrices $\mathbf{J}_k$ and $\mathbf{H}_k$ respectively. We consider row-orthonormalized matrices for compressive sensing, hence $\Phi\Phi^T$ is an identity matrix.

For LiSens case, similar to SPC, we have 
\begin{align}
\label{eq:proj_lisens}
    \mathbf{J_k} = \mathbf{H}_k-\Phi^T\left(\Phi\Phi^T\right)^{-1}\left(\Phi\mathbf{H}_k-\Y\right).
\end{align}

\subsection{Augmented Lagrangian method}
\label{sec:augmented_lag}
For the case of FlatCam reconstruction, the matrices $\mathbf{\Phi_L}\mathbf{\Phi_L}^T$ and $\mathbf{\Phi_R}\mathbf{\Phi_R}^T$ are ill-conditioned and can't be inverted. A closed form solution for projection operator doesn't exist. So, we consider the augmented Lagrangian corresponding to $\mathcal{C}_1$, with a dual parameter $\mathbf{\lambda}$.
\begin{multline}
\mathcal{L}(\X,\bm{\lambda}) = -log(p_{\theta}(\X)) + \rho\|\Y - \Phi_L \X \Phi_R^T \|^2_F \\ + \langle \bm{\lambda}, \Y - \Phi_L \X \Phi_R^T  \rangle_F
\end{multline}
However, instead of minimizing the Lagrangian with respect to the primal variable in each iteration, we just take one step of gradient descent. We further separate the gradient descent into two steps, one entirely depends on the prior while the other entirely depends on the imaging model. The update steps are as follows.
\begin{align}
    \mathbf{H}_k &= \X_k + \alpha \nabla_{\X}{log(p_{\theta}(\X_k)} \\
    \label{eq:proj_flatcam}
    \mathbf{J}_k &= \mathbf{H}_k + \Phi_L^T (\bm{\lambda}_k -  \rho(\Y - \Phi_L \X_k \Phi_R^T)) \Phi_R \\
    \X_{k+1} &= \Pi_{\mathcal{C}_2}(\mathbf{J}_{k}) \\
    \label{eq:dual_v}
    \bm{\lambda}_{k+1} &= \bm{\lambda}_k + \rho(\Y - \Phi_L \X_k \Phi_R^T )
\end{align}

\subsection{Soft constraint method}
\label{sec:soft_con}
Consider the case when the sensor has measurement noise,
\begin{align}
    \Y = f(\X) + \bm{\eta}
\end{align}
Assume the measurement noise $\bm{\eta}$ to be Gaussian distributed, i.e. 
\begin{align}
    \bm{\eta} &\sim \mathcal{N}(0,\sigma)\\
    \Y &\sim \mathcal{N}(f(\X),\sigma)
\end{align}
The MAP estimation problem can hence be reduced to 
\begin{align}
    \label{eq:proj_soft}
    \mathbf{\hat{X}} &= \argmax_{\X} ( log(p_{\theta}(\X)) + \lambda\| \Y - f(\X) \|^2 )
\end{align}
where $\lambda$ has to be estimated if we do not know the standard deviation of the measurement noise. Since the constraints are not exact here, we replace the step to project to the constraint space by instead taking a step towards minimizing the likelihood. Hence, we replace Eq.\ \ref{eq:project} by gradient descent over likelihood, 
\begin{align}
    \mathbf{J}_k = \mathbf{H}_k - \alpha f'(\mathbf{H}_k) (\Y - f(\mathbf{H}_k))
\end{align}


\subsection{Stochastic gradients using pixel dropout}
\label{sec:stoch_grad}
We observe that if we update all the pixels in the gradient update (Eq.\ \ref{eq:ascent}), then we get washed out reconstructions. The autoregressive prior directly models correlation between neighbouring pixels. Hence it tends to assign same values to neighbouring problems. We combat this problem by randomly selecting a certain amount of pixels to update in each step. Hence, not all pixels get updates at every step. We call this pixel dropout, and for incorporating that, we replace the gradient in Eq.\ \ref{eq:ascent} by stochastic gradients, i.e.,
\begin{align}
\label{eq:dropout}
    \mathbf{H}_k &= \X_{k} + \alpha \mathbf{M} \circ \nabla_{\X} log(p_{\theta}(\X_k))
\end{align}
where  $\mathbf{M}$ is a random binary mask with the percentage of zeros determined by the pixel dropout ratio. This is analogous to the case of training deep neural networks, where Stochastic Gradient Descent (SGD) helps in escaping from sharp local minima \cite{keskar2016large}. Here, the washed out reconstructions correspond to sharp local minima owing to the strong correlation between pixels. We demonstrate the effect of the amount of pixel dropout on the reconstructions in Section \ref{sec:abalation}. 

\subsection{Splitting and Stitching}
\label{sec:splitting}
Our prior model is trained on $64 \times 64$ patches, hence the input for $p_{\theta}(\X)$ has to be $64 \times 64$. While we perform the likelihood step on the entire image, our approach is designed such that the prior gradient update, projection, and clipping steps are separate. Before the prior gradient update, we split the image into a batch of $64 \times 64$ patches. Before performing the likelihood step, we stitch the patches back into original dimensions.  

Our approach is summarized as follows:

\begin{algorithm}[h]
\SetAlgoLined
\KwData{Simulated or real measurements $\Y$, Simulated or calibrated imaging matrix $\Phi$,
Learned autoregressive prior model $p_{\theta}(\X)$}
\KwResult{Reconstructed $N\times N$ image $\X$}
Initialization: $\X_{ij} \sim \mathcal{U}(0,1)$ $\forall$ pixels $i,j$
\While{iterations $<$ max\_iter}{
    Split $\X$ into a batch of $64\times64$ patches
    \For{Gradient ascent w.r.t $p_{\theta}(\X)$}{Obtain  $\nabla_{\X}{p_{\theta}(\X)}$ via back-prop to inputs \\ 
    Apply pixel dropout mask and update $\X$ (Eq.\ \ref{eq:dropout})}
    Stitch $\X$ back into $N\times N$ image\\
    \For{Satisfying constraints}{
    Clip $\X_{ij}$ between $0$ and $1$ $\forall$ $i,j$\\
    Project the solution to the constraint space specified by the forward model and inference method appropriately (Eq.\ \ref{eq:proj_inpaint},\ref{eq:proj_spc},\ref{eq:proj_lisens},\ref{eq:proj_flatcam} or \ref{eq:proj_soft})}
    \If{method is augmented Lagrangian}{
    Update dual variable $\bm{\lambda}$ (Eq.\ \ref{eq:dual_v})
    }
}
\caption{Our image reconstruction algorithm}
\end{algorithm}

\begin{figure*}[t] 
    \centering
\begin{minipage}{\textwidth}
\centering
\begin{minipage}{.24\textwidth}
\centerline{Original image}
\vspace{0.01cm}
\includegraphics[trim=2 2 2 2,clip, width=1\textwidth]{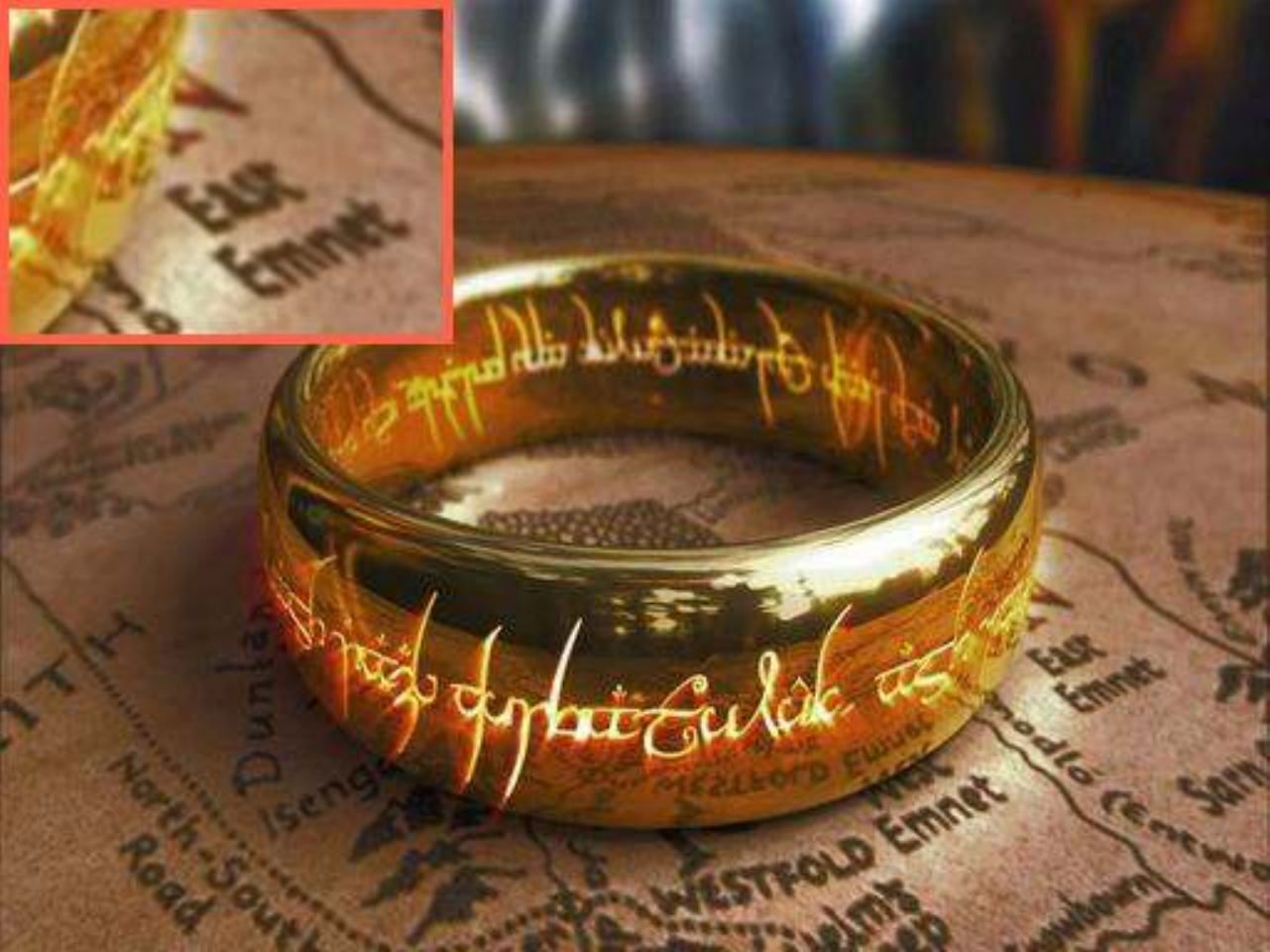}  
\centerline{}
\end{minipage}
\begin{minipage}{.24\textwidth}
\centerline{Masked image}
\vspace{0.1cm}
\includegraphics[trim=2 2 2 2,clip, width=1\textwidth]{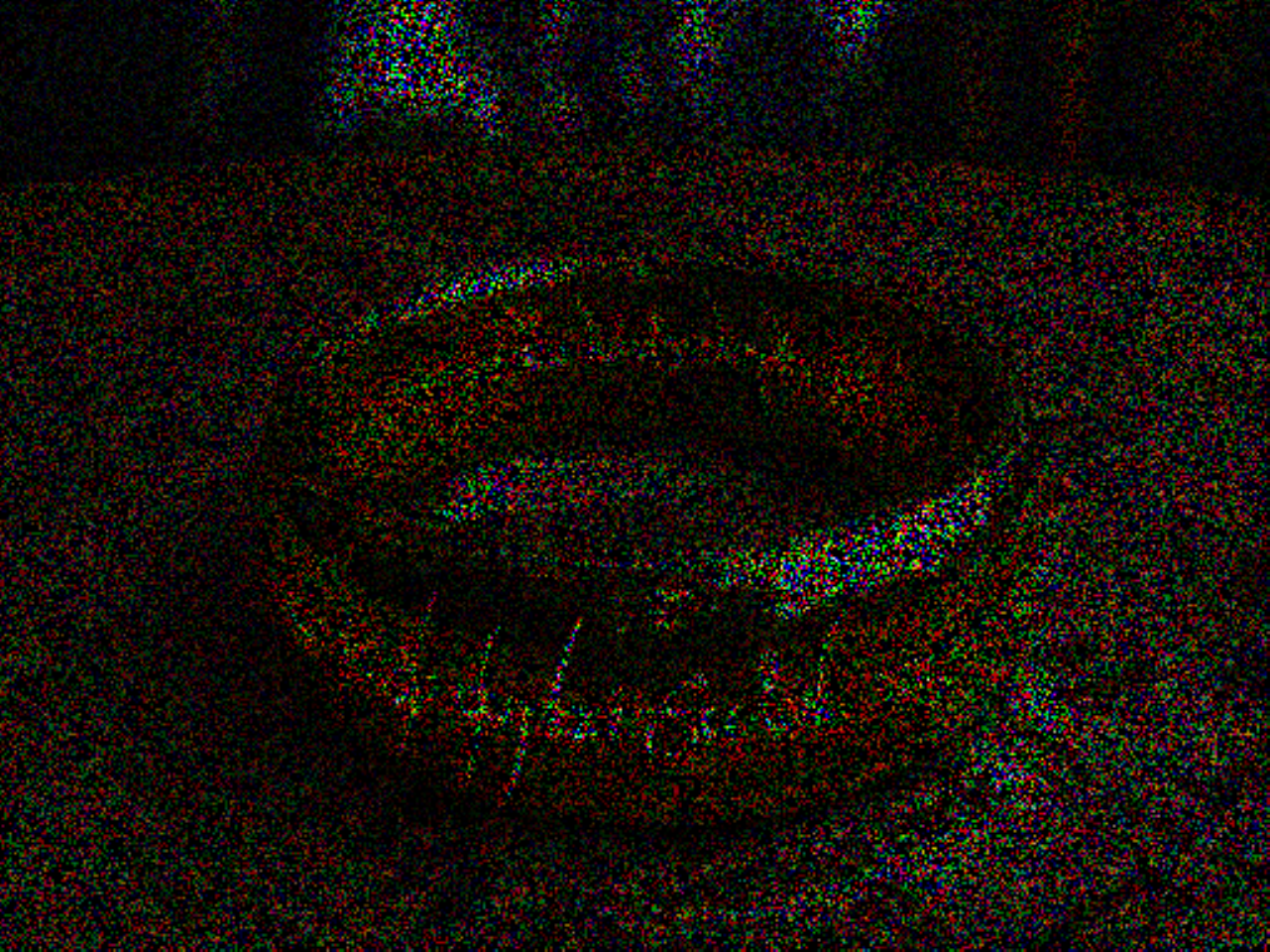}  
  
\centerline{ 80\% missing pixels}
\end{minipage}
\begin{minipage}{.24\textwidth}
\centerline{OneNet}
\vspace{0.1cm}
\includegraphics[trim=2 2 2 2,clip, width=1\textwidth]{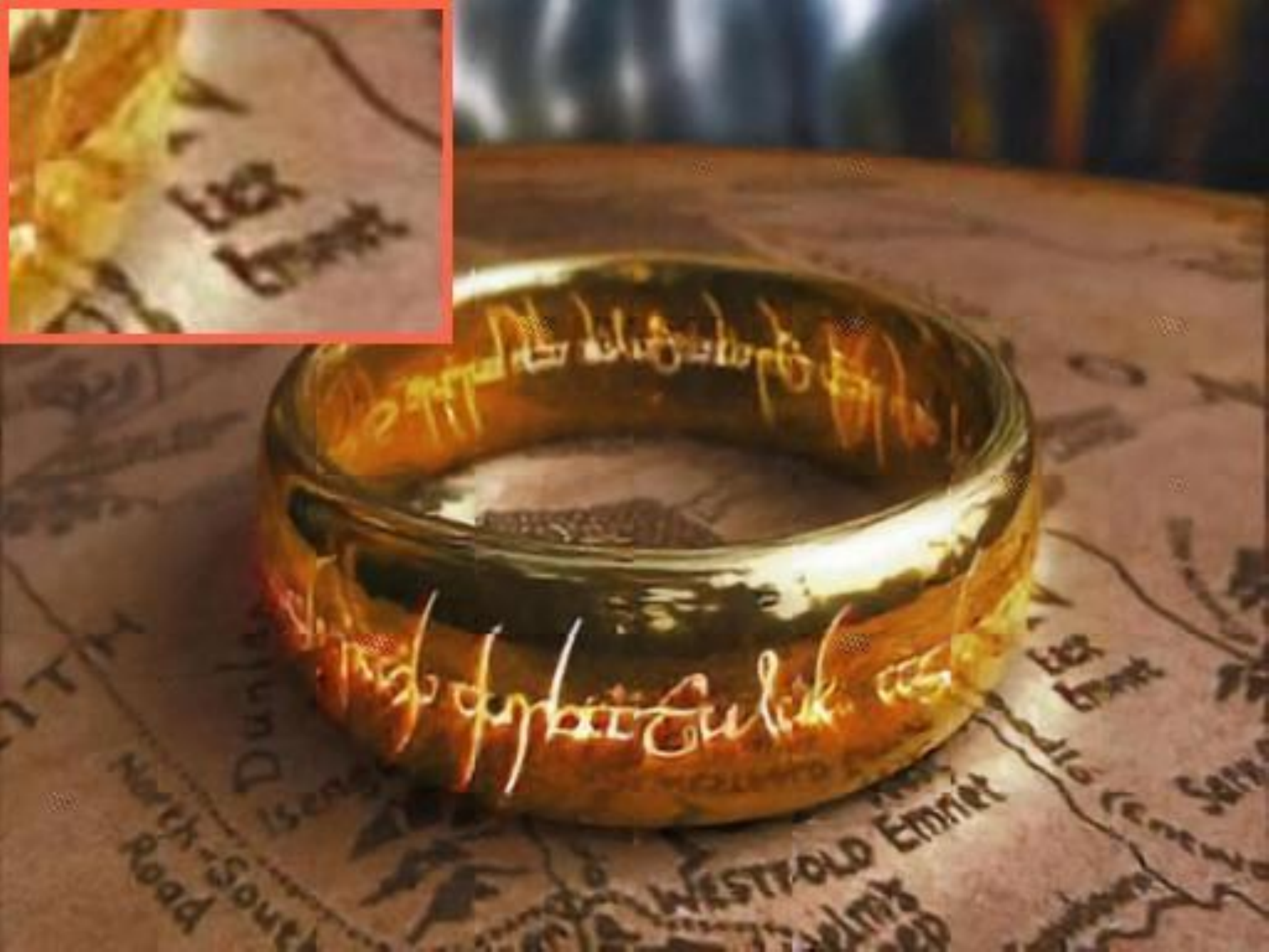}  

\centerline{ 30.71 dB, 0.909}
\end{minipage}
\begin{minipage}{.24\textwidth}
\centerline{Ours}
\vspace{0.1cm}
\includegraphics[trim=2 2 2 2,clip, width=1\textwidth]{inpainting/ring_full/mr_20/orig.pdf}  
\centerline{ 33.61 dB, 0.971}

\end{minipage}\\

\end{minipage}
    \caption{Random pixel inpainting with 80\% missing pixels. Our approach reconstructs the finer edges better and has more consistency among neighbouring pixels, as compared to OneNet. Note the details around the text shown in zoomed patch. The difference between the two reconstructions can be perceived by further zooming into the images. The numbers reported in this and the subsequent figures are PSNR (in dB) followed by SSIM. }
    \label{fig:inpainting}
\end{figure*}

\section{Implementation Details}

\subsection{Our Approach}
\label{sec:our_implementation}
We train PixelCNN++ on the downsampled $64\times64$ ImageNet data as introduced in \cite{van2016pixel} for 6 epochs. Batch size is kept as 36 and the number of filter channels as 100. The rest of the parameters are same as the ones used for training PixelCNN++ on $32\times32$ ImageNet in \cite{salimans2017pixelcnn++}. We obtain a negative log likelihood score of 3.66 on test data and 3.5 on train data which is consistent with the numbers reported in \cite{salimans2017pixelcnn++} for similar data. 

With this learned model, we use our proposed algorithm as described in Algorithm 1, for the experiments described below. An initial image is sampled from a uniform random distribution. However, we observe that starting with different initial images doesn't have much effect on the final converged reconstruction. We use momentum in the gradient update for faster convergence, with its value set to 0.9. Step size $\alpha$, maximum iterations, likelihood weightage $\rho$ for each experiment are mentioned in the subsequent section.

For reconstructing color images, we consider multiplexing along individual color channels. Hence, we have separate $\Phi$ matrices for all the three channels and obtain three separate measurement vectors $\Y$ for each channel.

We have made the code of our implementation for the task Single Pixel Camera reconstruction available online\footnote{https://github.com/adaveiitm/deep-pixel-level-prior}. 

\subsection{One Network to solve them all}
\label{sec:one_net_change}
We use the original implementation of \cite{chang2017one} available online\footnote{https://github.com/rick-chang/OneNet} with certain modification as mentioned below. 

For simulating color Single Pixel Camera, the original implementation rasterizes the entire $N \times N \times 3$ image into a single vector and creates one $\Phi$ matrix to compress this into a single measurement vector. We believe that this might not be feasible to implement in a real system. Hence, we modify their implementation to instead simulate separate $\Phi$ matrices for each channel as in Section \ref{sec:our_implementation}.  

While simulating SPC measurements on large images, the original implementation only deals with local multiplexing. It breaks them down into patches of $64 \times 64$ and compresses each of these patches separately. We modify this to deal with the more challenging case of global multiplexing, where we compress the entire image. 

We extend the original implementation for LiSens and FlatCam as well, by considering the above modifications and incorporating the respective forward models. 

We use model provided which was trained on $64 \times 64$ Imagenet for 2 epochs for testing the results. We found that the results were very much dependent on the alpha parameter (penalty parameter) which had to be tuned for each image to get the best solution. 

\subsection{TVAL3}
For comparisons with TVAL3 ( TV minimization by Augmented Lagrangian and ALternating direction algorithms ) \cite{li2013efficient}, we use the MATLAB implementation\footnote{http://www.caam.rice.edu/~optimization/L1/TVAL3/} with the default parameters. The number of iterations is set to 80. For color image reconstruction, we update each channel separately using TVAL3.

\section{Experiments}
In this section, we present the reconstructions from our approach and compare them with the existing state-of-the-art approaches. To being with, we illustrate the ability of an autoregressive prior in reconstructing pixel level details using an example of missing pixel inpainting in an image. For this, we randomly mask out pixels from the image and use our prior to reconstruct these missing pixels. We perform by keeping the observed pixel values as same and update missing pixels to maximize the the prior loglikelihood. Specifically, we take an image of size 384x512 and mask 80\% of the pixels in the initial image as could be seen in Figure \ref{fig:inpainting}. We compare our results with that of OneNet \cite{chang2017one}, and we can observe details in our reconstruction much better like the text outlines, also quantitatively in terms of PSNR and SSIM. We use a step size of 75 and run for approximately 1000 iterations. 

For all the three imaging setups of SPC, Lisens and Flatcam we perform reconstructions on both simulated data and real measurements. In case of simulation we compare our reconstructions with TVAL3 \cite{li2013efficient} and OneNet \cite{chang2017one}. In case of reconstructions from real measurements, we compare our results with TVAL3. OneNet experiments failed to converge to a stable point in this case hence we could not provide comparison with this approach. For real Lisens at 66\% measurements, although OneNet converges, results obtained were very poor compared to other approaches.

\subsection{Single Pixel Camera}

\subsubsection{Simulation case}
\begin{figure*}[t]
\begin{minipage}{.01\textwidth}
\raggedleft
\begin{turn}{90}Reconstructions with 5\% M.R. \end{turn}
\end{minipage}
\begin{minipage}{.98\textwidth}
\centering
\begin{minipage}{.20\textwidth}
\centerline{Original image}
\vspace{0.01cm}
\includegraphics[trim=2 2 2 2,clip, width=1\textwidth]{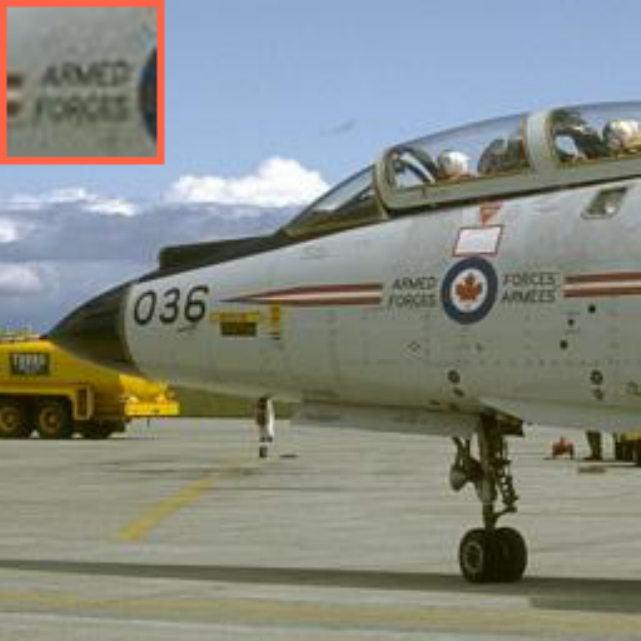}  
\centerline{}
\end{minipage}\hspace{0.1cm}
\begin{minipage}{.20\textwidth}
\centerline{TVAL3}
\vspace{0.01cm}
\includegraphics[trim=2 2 2 2,clip, width=1\textwidth]{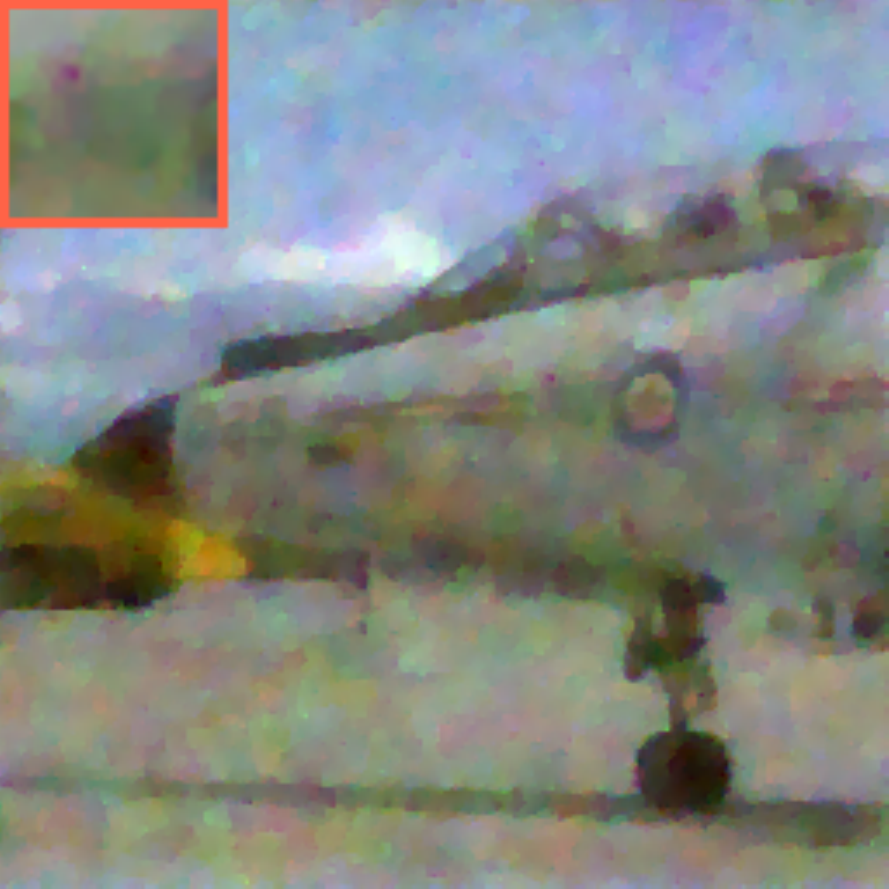}  
\centerline{ 22.54 dB, 0.705}
\end{minipage}\hspace{0.1cm}
\begin{minipage}{.20\textwidth}
\centerline{OneNet}
\vspace{0.1cm}
\includegraphics[trim=2 2 2 2,clip, width=1\textwidth]{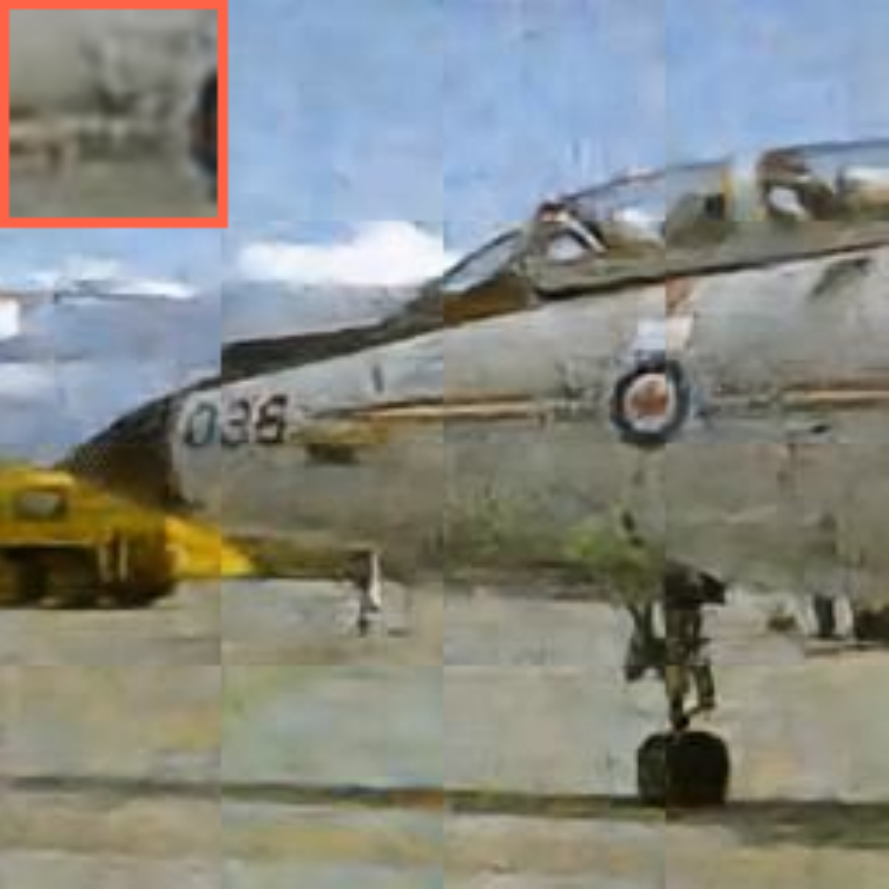}  
\centerline{ 26.81 dB, 0.861}
\end{minipage}\hspace{0.1cm}
\begin{minipage}{.20\textwidth}
\centerline{Ours}
\vspace{0.1cm}
\includegraphics[trim=2 2 2 2,clip, width=1\textwidth]{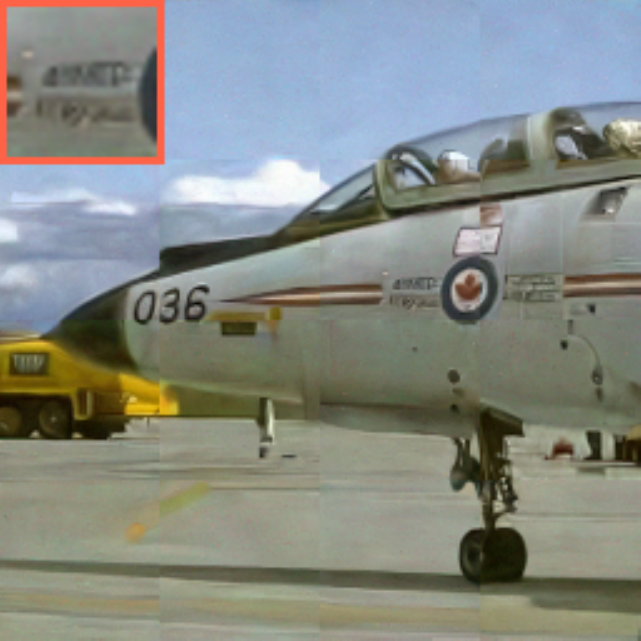}  
\centerline{ 30.51 dB, 0.918}
\end{minipage}\\

\begin{minipage}{.20\textwidth}
\vspace{0.04cm}
\includegraphics[trim=2 2 2 2,clip, width=1\textwidth]{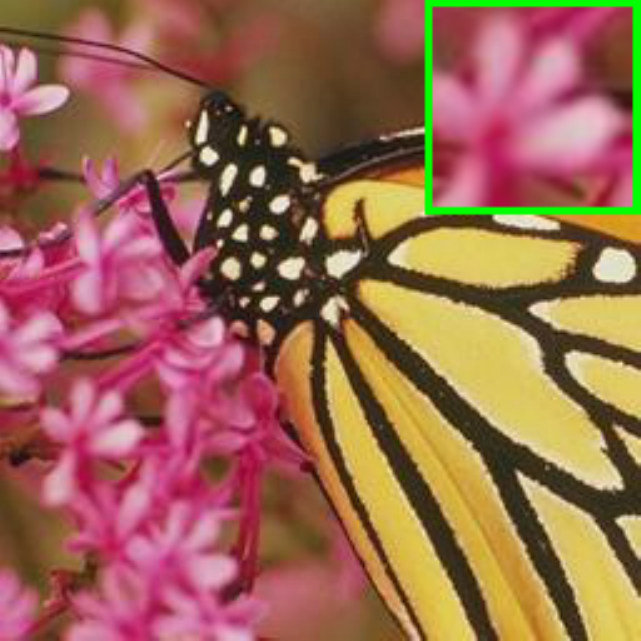}  
\centerline{}
\end{minipage}\hspace{0.1cm}
\begin{minipage}{.20\textwidth}
\vspace{0.04cm}
\includegraphics[trim=2 2 2 2,clip, width=1\textwidth]{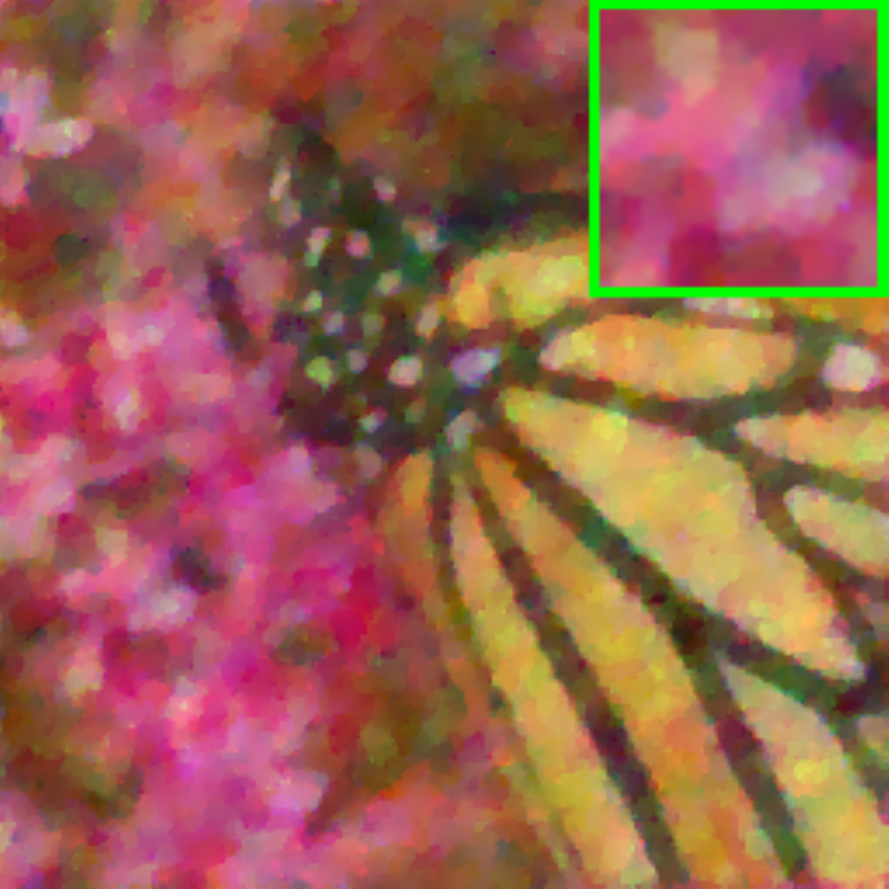}  
\centerline{ 18.73 dB, 0.823}
\end{minipage}\hspace{0.1cm}
\begin{minipage}{.20\textwidth}
\vspace{0.1cm}
\includegraphics[trim=2 2 2 2,clip, width=1\textwidth]{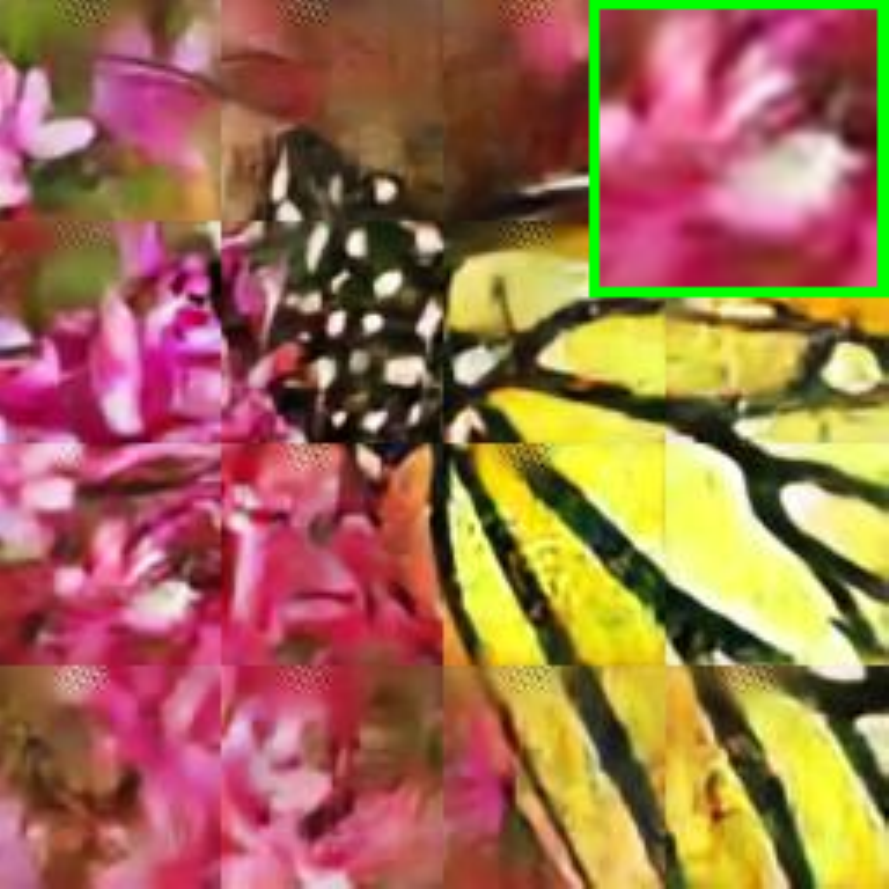}  
\centerline{ 19.50 dB, 0.874}
\end{minipage}\hspace{0.1cm}
\begin{minipage}{.20\textwidth}
\vspace{0.1cm}
\includegraphics[trim=2 2 2 2,clip, width=1\textwidth]{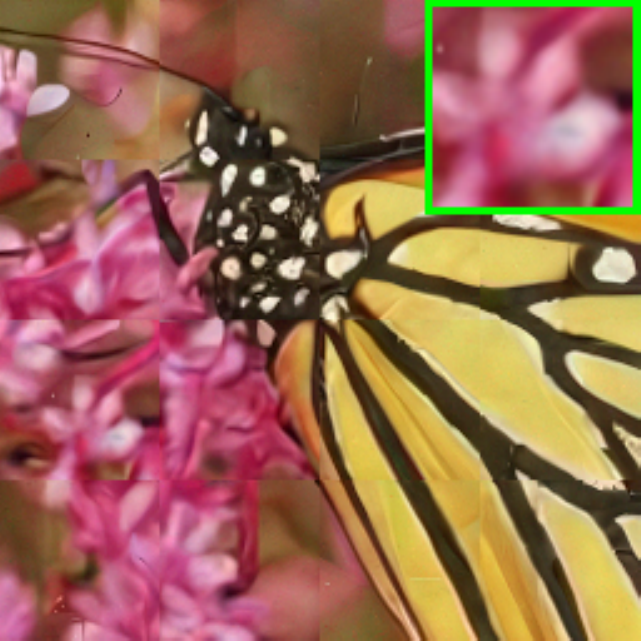}  
\centerline{ 26.28 dB, 0.954}
\end{minipage}
\end{minipage}

\centerline{}
\centerline{}

\begin{minipage}{.01\textwidth}
\raggedleft
\begin{turn}{90}Reconstructions with 10\% M.R. \end{turn}
\end{minipage}
\begin{minipage}{.98\textwidth}
\centering
\begin{minipage}{.20\textwidth}
\vspace{0.01cm}
\includegraphics[trim=2 2 2 2,clip, width=1\textwidth]{spc/color/plane/mr_05/orig.pdf}  
\centerline{}
\end{minipage}\hspace{0.1cm}
\begin{minipage}{.20\textwidth}
\vspace{0.01cm}
\includegraphics[trim=2 2 2 2,clip, width=1\textwidth]{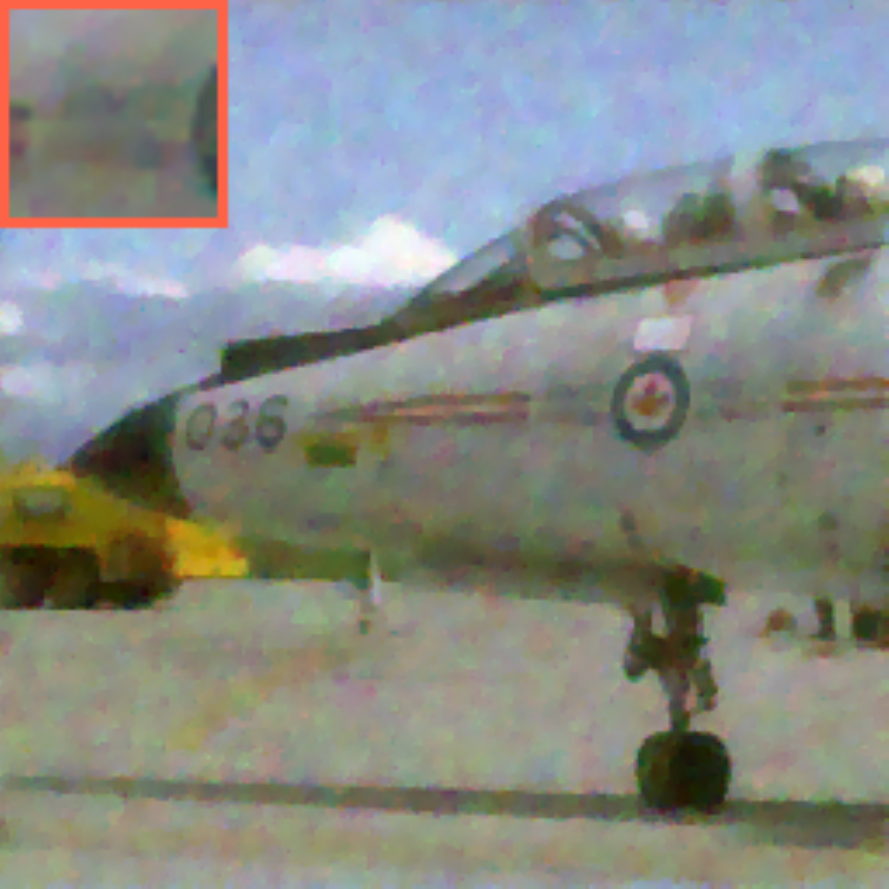}  
\centerline{ 25.46 dB, 0.815}
\end{minipage}\hspace{0.1cm}
\begin{minipage}{.20\textwidth}
\vspace{0.1cm}
\includegraphics[trim=2 2 2 2,clip, width=1\textwidth]{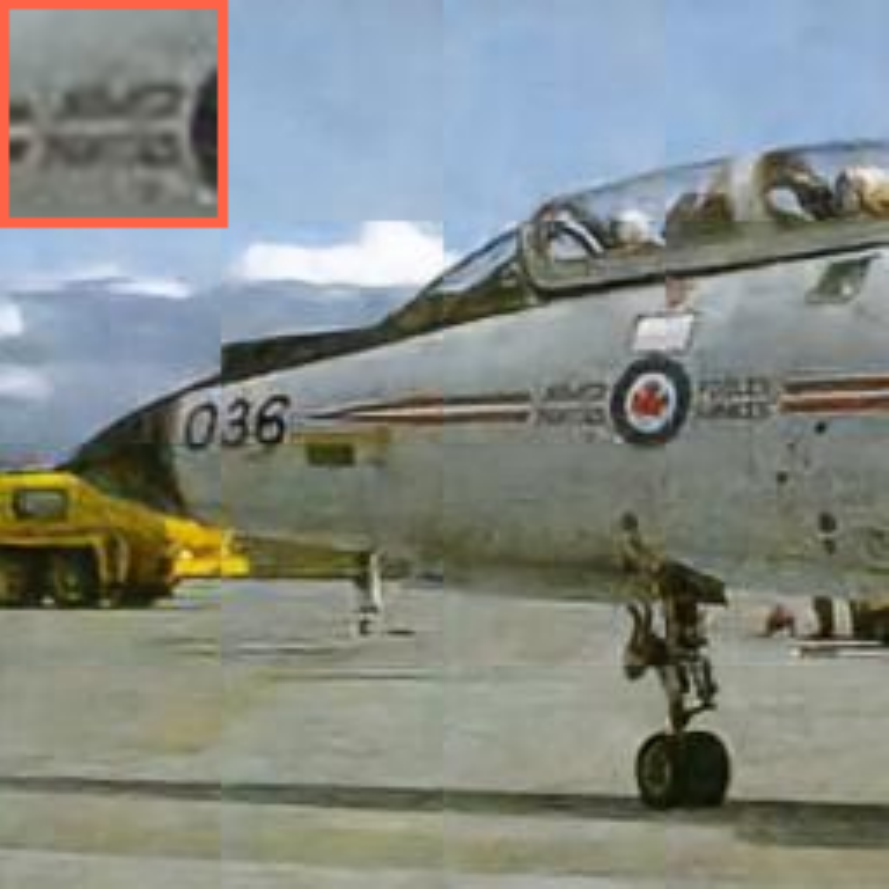}  
\centerline{28.72 dB, 0.897}
\end{minipage}\hspace{0.1cm}
\begin{minipage}{.20\textwidth}
\vspace{0.1cm}
\includegraphics[trim=2 2 2 2,clip, width=1\textwidth]{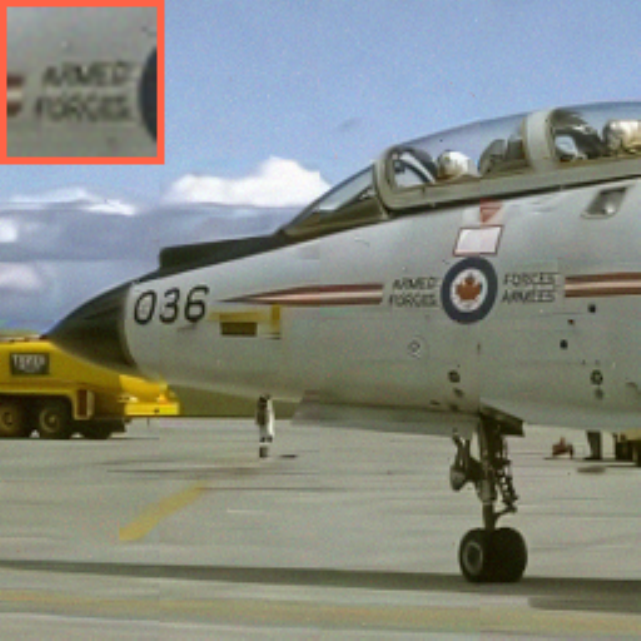}  
\centerline{ 32.46 dB, 0.935}
\end{minipage}

\begin{minipage}{.20\textwidth}
\vspace{-0.35cm}
\includegraphics[trim=2 2 2 2,clip, width=1\textwidth]{spc/color/monarch/mr_05/orig.pdf}  
\end{minipage}\hspace{0.1cm}
\begin{minipage}{.20\textwidth}
\vspace{0.1cm}
\includegraphics[trim=2 2 2 2,clip, width=1\textwidth]{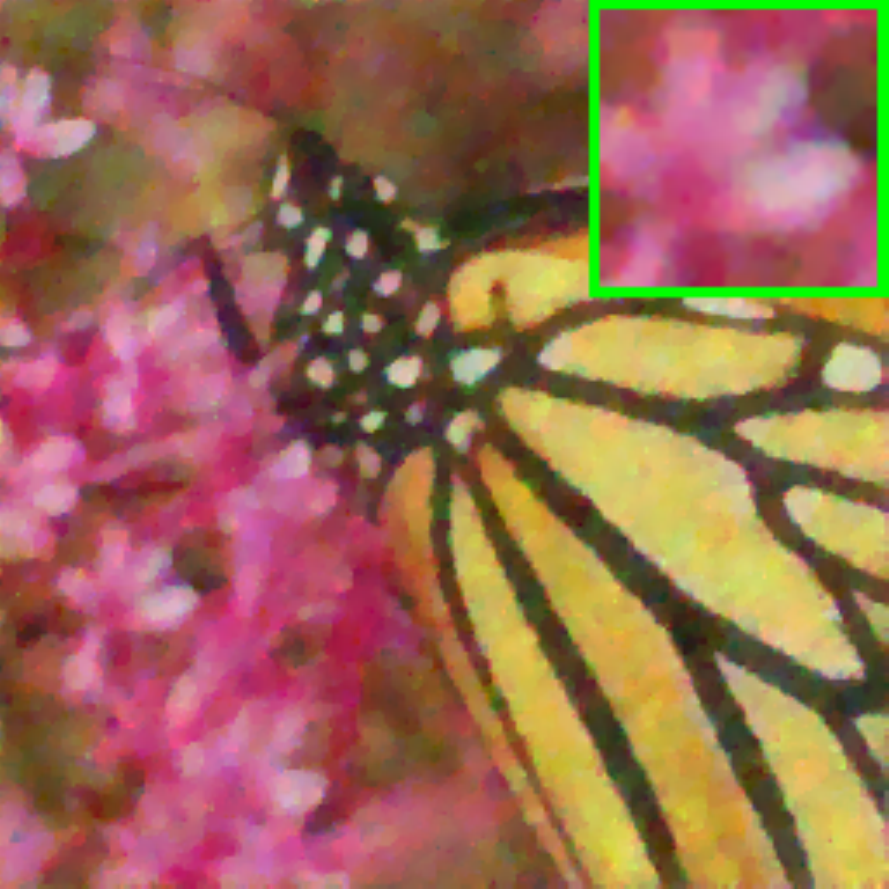}  
\centerline{ 21.83 dB, 0.893}
\end{minipage}\hspace{0.1cm}
\begin{minipage}{.20\textwidth}
\vspace{0.1cm}
\includegraphics[trim=2 2 2 2,clip, width=1\textwidth]{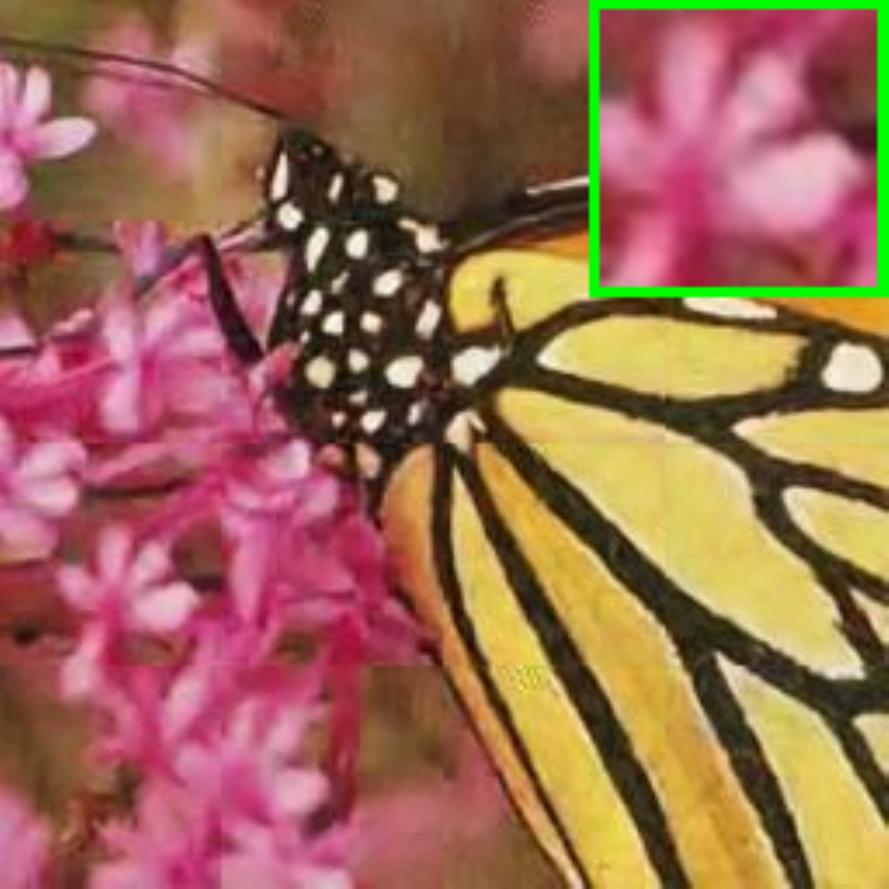}  
\centerline{27.58 dB, 0.965}
\end{minipage}\hspace{0.1cm}
\begin{minipage}{.20\textwidth}
\vspace{0.1cm}
\includegraphics[trim=2 2 2 2,clip, width=1\textwidth]{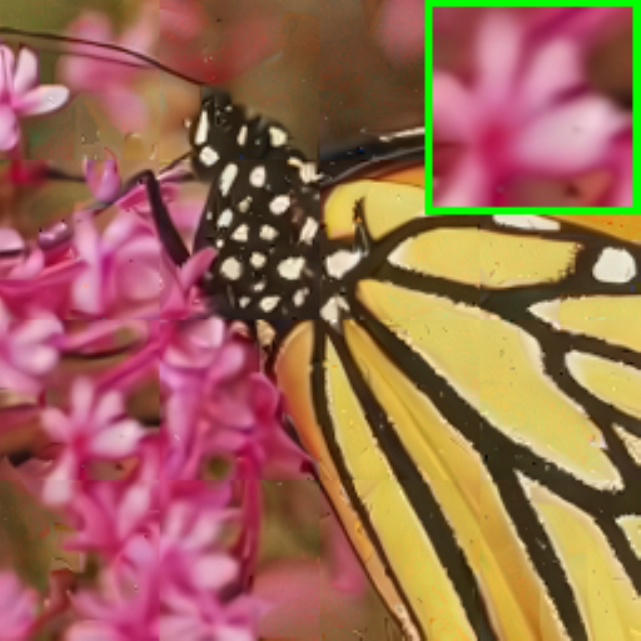}  
\centerline{29.69 dB, 0.976}
\end{minipage}
\centerline{}
\centerline{}
\end{minipage}
\caption{Qualitative comparisons of $256 \times 256$ images reconstructed from simulated Single Pixel Camera measurements using TVAL3, OneNet and our approach. Even when the measurement rate is low, our method reconstructs the sharp and promiment structures in the image better. Moreover, there are no visible artifacts in our reconstructions as the autoregressive prior ensures the nearby pixels to be consistent. This is not the case with TVAL3 and OneNet leadning to poor performance.}
\label{fig:spc_color}
\end{figure*}

We show quantitative and qualitative comparisons of simulated SPC reconstruction results on images of sizes 128$\times$128 and 256$\times$256 respectively as shown in Table \ref{table:spc} and Figure \ref{fig:spc_color} respectively. Measurement rates considered are $10\%$ and $25\%$ for 128$\times$128 and 5\% and 10\% for 256$\times$256. Similar to RIDE-CS \cite{dave2017compressive}, we generate the $\phi$ matrix as a random Gaussian with orthonormal rows. We perform gradient descent and projection operation on the compressed image for 2000 iterations in the case of 25\% measurement rate and for 2500 iterations in case of 10\% measurement rate. We use a step-size of 7.5 and the hard constraint projection method. In all cases, we intialize with random image from uniform distribution.  We compare our results to \cite{chang2017one} and we are able to show significant improvement in reconstruction results in terms of PSNR and SSIM values. Our reconstructions have better edges and textures compared to the reconstructions from OneNet.

\begin{figure}[t]
    \centering
    \subfloat[bird]{\includegraphics[width=0.6in]{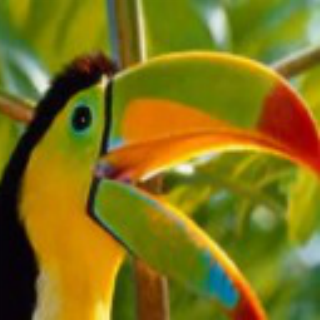}}
    \hspace{0.02 in}
    \subfloat[building]{\includegraphics[width=0.6in]{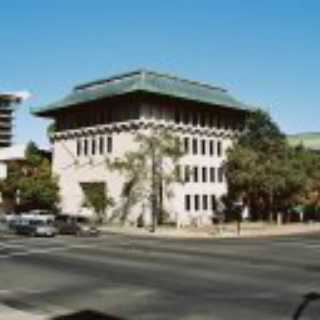}}
    \hspace{0.02 in}
    \subfloat[cat]{\includegraphics[width=0.6in]{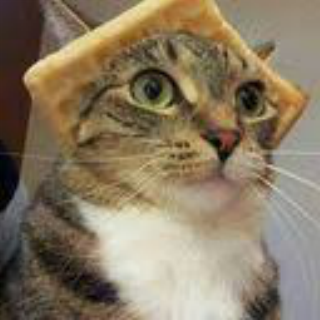}}
    \hspace{0.02 in}
    \subfloat[flower]{\includegraphics[width=0.6in]{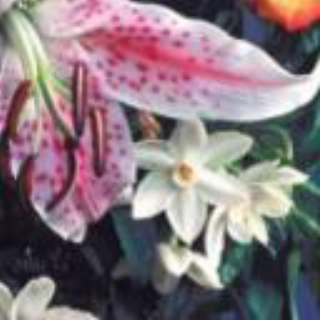}}
    \hspace{0.02 in}
    \subfloat[parrot]{\includegraphics[width=0.6in]{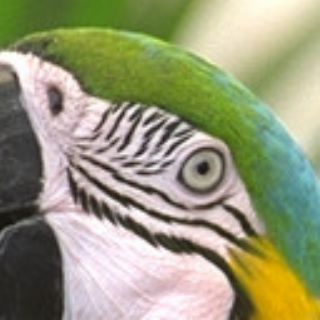}}
    \caption{Test images of $128 \times 128$ size chosen randomly for simulated SPC and LiSens reconstructions.}
    \label{fig:sel_imgs}
\end{figure}
\subsubsection{Real Case}

\begin{table}[h]
\renewcommand{\arraystretch}{1.3}
\caption{Comparisons of reconstructions from simulated SPC measurements at different measurement rates for the images shown in Figure \ref{fig:sel_imgs}. Our approach obtains better performance than OneNet and TVAL3 by modelling pixel-level consistencies. See Figure \ref{fig:spc_color} for qualitative comparisons}
\label{table:spc}
\begin{center}
\begin{tabular}{cccccccc}

\toprule
    \multirow{2}{*}{ Name}&\multirow{2}{*}{M.R.}&\multicolumn{2}{c}{TVAL3}&\multicolumn{2}{c}{OneNet}&\multicolumn{2}{c}{Ours}\\
    \cline{3-8} & &  PSNR & SSIM & PSNR & SSIM  & PSNR & SSIM  \\
\midrule
 \multirow{2}{*}{bird} &10 & 23.67 & 0.91 & 23.92 & 0.93 & \textbf{29.52} & \textbf{0.97} \\
&25 & 29.67 & 0.97 & 26.89 & 0.96 & \textbf{32.96} & \textbf{0.98} \\
\midrule
 \multirow{2}{*}{building} &10 & 18.81 & 0.61 & 23.85 & 0.86 & \textbf{25.93} & \textbf{0.88} \\
&25  & 22.72 & 0.79 & 24.06 & 0.87 & \textbf{32.05} & \textbf{0.96} \\
\midrule
 \multirow{2}{*}{cat} &10  & 23.27 & 0.72 & 25.15 & 0.82 & \textbf{26.68} & \textbf{0.85} \\
&25  & 26.87 & 0.85 & 26.60 & 0.88 & \textbf{31.23} & \textbf{0.94} \\
\midrule
 \multirow{2}{*}{flower} &10  & 20.07 & 0.68 & 23.39 & 0.84 & \textbf{26.22} & \textbf{0.89} \\
&25  & 24.84 & 0.86 & 25.13 & 0.90 & \textbf{31.05} & \textbf{0.96} \\
\midrule
 \multirow{2}{*}{parrot} &10  & 18.49 & 0.64 & 25.82 & 0.89 & \textbf{27.59} & \textbf{0.90} \\
&25  & 23.67 & 0.84 & 26.79 & 0.91 & \textbf{32.18} & \textbf{0.95} \\
\midrule
 \multirow{2}{*}{mean} &10  &  20.86 & 0.72 & 24.43 & 0.87 & \textbf{27.19} & \textbf{0.90} \\
&25  &  25.55 & 0.86 & 25.74 & 0.90 & \textbf{31.89} & \textbf{0.96} \\
\bottomrule
\end{tabular}
\end{center}
\end{table}

\begin{figure*}[!t]
\begin{minipage}{.001\textwidth}
\raggedleft
\begin{turn}{90} 15\% M.R. \end{turn}
\end{minipage}
\begin{minipage}{.999\textwidth}
\centering
\begin{minipage}{.20\textwidth}
\centerline{Original image}
\vspace{0.01cm}
\includegraphics[trim=2 2 2 2,clip, width=1\textwidth]{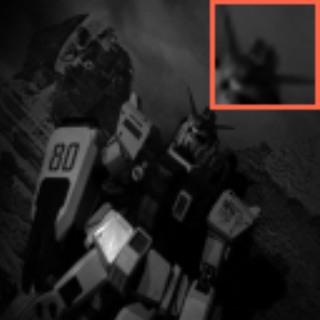}  
\centerline{}
\end{minipage}\hspace{0.1cm}
\begin{minipage}{.20\textwidth}
\centerline{TVAL3}
\vspace{0.1cm}
\includegraphics[trim=2 2 2 2,clip, width=1\textwidth]{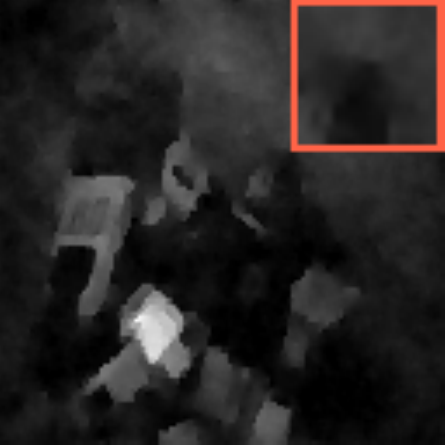}  
  
\centerline{28.24 dB, 0.862}
\end{minipage}\hspace{0.1cm}
\begin{minipage}{.20\textwidth}
\centerline{RIDE-CS}
\vspace{0.1cm}
\includegraphics[trim=2 2 2 2,clip, width=1\textwidth]{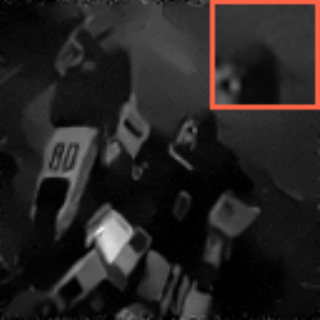}  
  
\centerline{30.70 dB, 0.910}
\end{minipage}\hspace{0.1cm}
\begin{minipage}{.20\textwidth}
\centerline{Ours}
\vspace{0.1cm}
\includegraphics[trim=2 2 2 2,clip, width=1\textwidth]{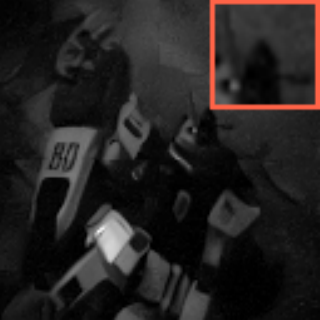}  
\centerline{31.65 dB, 0.913}
\end{minipage}\\

\end{minipage}

\centerline{}
\centerline{}

\begin{minipage}{.001\textwidth}
\raggedleft
\begin{turn}{90} 30\% M.R. \end{turn}
\end{minipage}
\begin{minipage}{.999\textwidth}
\centering
\begin{minipage}{.20\textwidth}
\centerline{Original image}
\vspace{0.01cm}
\includegraphics[trim=2 2 2 2,clip, width=1\textwidth]{spc/real/tf/mr_15/orig.pdf}  
\centerline{}
\end{minipage}\hspace{0.1cm}
\begin{minipage}{.20\textwidth}
\centerline{TVAL3}
\vspace{0.1cm}
\includegraphics[trim=2 2 2 2,clip, width=1\textwidth]{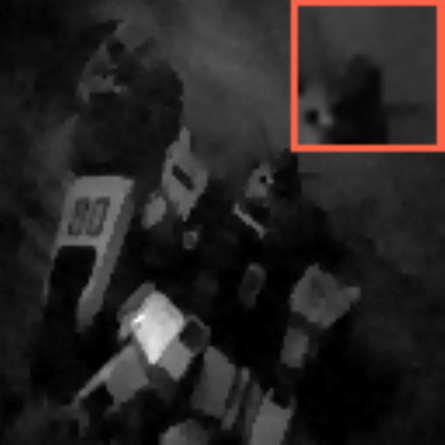}  
  
\centerline{32.17 dB, 0.922}
\end{minipage}\hspace{0.1cm}
\begin{minipage}{.20\textwidth}
\centerline{RIDE-CS}
\vspace{0.1cm}
\includegraphics[trim=2 2 2 2,clip, width=1\textwidth]{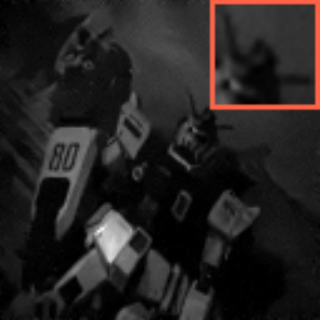}  
  
\centerline{36.96 dB, 0.972}
\end{minipage}\hspace{0.1cm}
\begin{minipage}{.20\textwidth}
\centerline{Ours}
\vspace{0.1cm}
\includegraphics[trim=2 2 2 2,clip, width=1\textwidth]{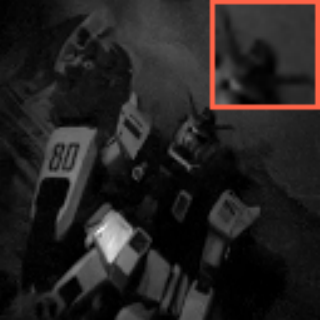}  
\centerline{36.85 dB, 0.970}
\end{minipage}\\

\end{minipage}

\caption{Reconstructions from real Single Pixel Camera measurements at different measurement rates. Our approach recovers the low level details much better than TVAL3. Though the performance of RIDE-CS, which is also a deep autoregressive model, is similar to ours in this case, its computational complexity is much higher. Also in other simulation experiments we found RIDE-CS does not preserve fine details, see Figure \ref{fig:abalation_ride}.}
\label{fig:spc_real}
\end{figure*}

We show our real SPC reconstruction results in Figure \ref{fig:spc_real}. Data for this experiment is provided to us by the authors of \cite{wang2015lisens}. We obtain the real SPC sensor measurements at 30\% and 15\% measurement rate respectively. The images we reconstruct in this case are grey scale images. Here also, we use the Hard constraint projection method for inference. We compare our results with TVAL3 and RIDE-CS \cite{dave2017compressive}. Our method performs better than both RIDE-CS and TVAL3 in terms of PSNR and SSIM values. Apart from these measures, we observe that our method produces a sharper reconstruction. We use the same hyperparameters and training procedures as in the simulated case. 

\subsection{LiSens}
\subsubsection{Simulation case}
\begin{figure*}[t]
\begin{minipage}{.01\textwidth}
\raggedleft
\begin{turn}{90}Reconstructions with 25\% M.R. \end{turn}
\end{minipage}
\begin{minipage}{.98\textwidth}
\centering
\begin{minipage}{.20\textwidth}
\centerline{Original image}
\vspace{0.01cm}
\includegraphics[trim=2 2 2 2,clip, width=1\textwidth]{spc/color/plane/mr_05/orig.pdf}  
\centerline{}
\end{minipage}\hspace{0.1cm}
\begin{minipage}{.20\textwidth}
\centerline{TVAL3}
\vspace{0.01cm}
\includegraphics[trim=2 2 2 2,clip, width=1\textwidth]{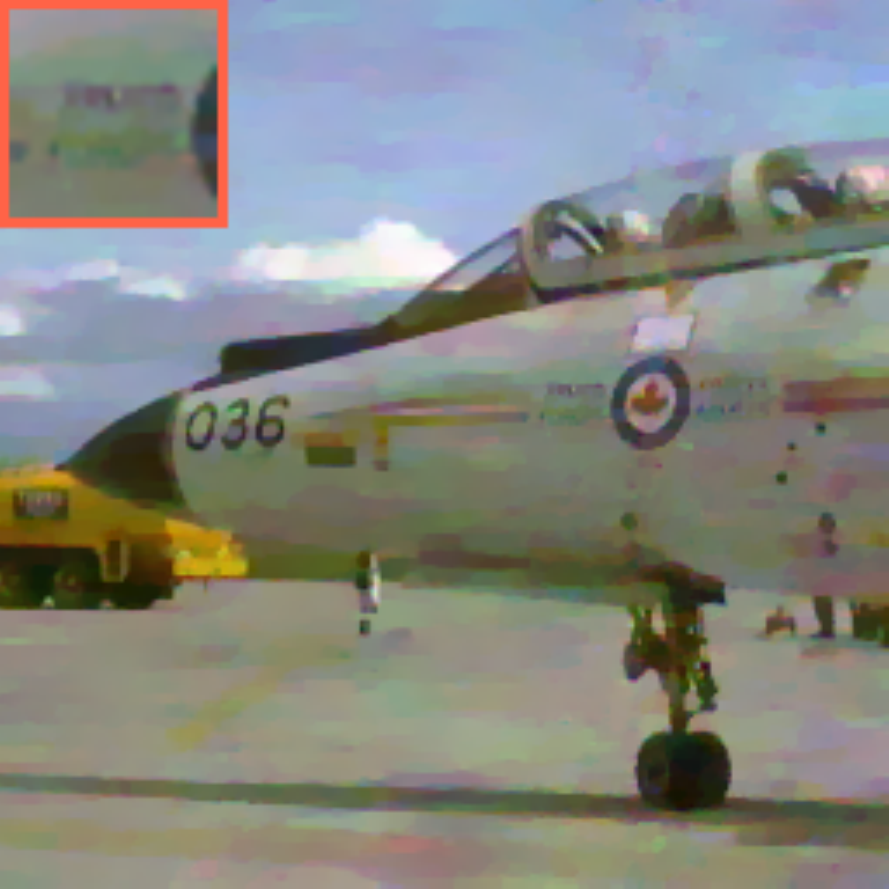}  
\centerline{ 26.43 dB, 0.846}
\end{minipage}\hspace{0.1cm}
\begin{minipage}{.20\textwidth}
\centerline{OneNet}
\vspace{0.1cm}
\includegraphics[trim=2 2 2 2,clip, width=1\textwidth]{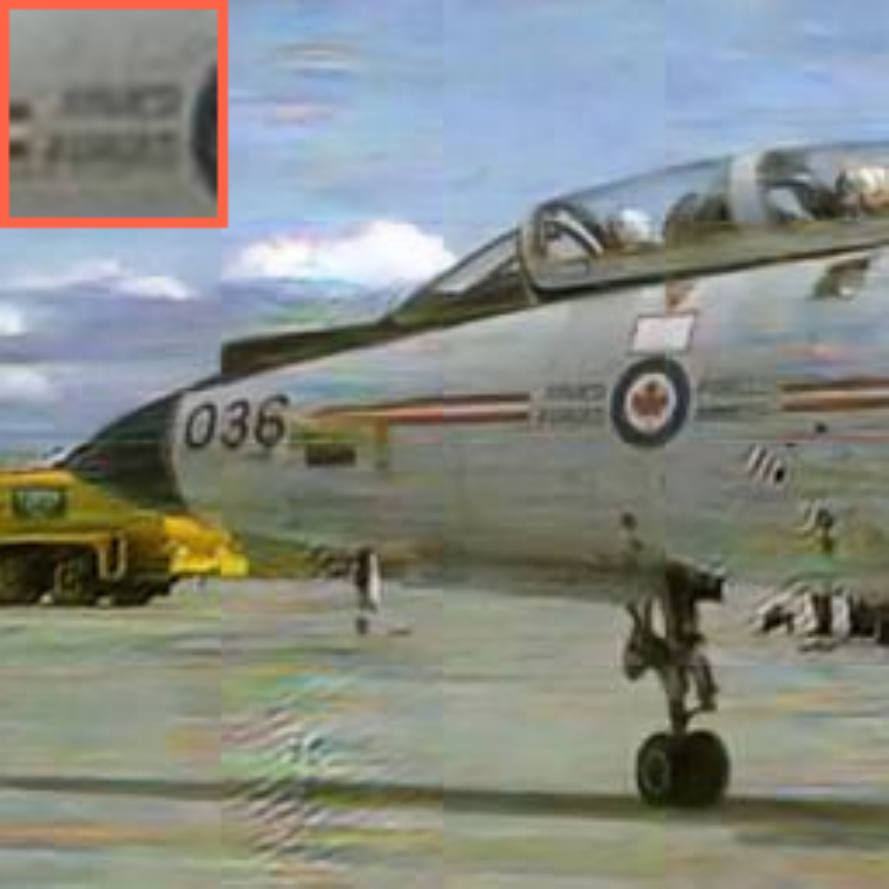}  
  
\centerline{ 26.12 dB, 0.845}
\end{minipage}\hspace{0.1cm}
\begin{minipage}{.20\textwidth}
\centerline{Ours}
\vspace{0.1cm}
\includegraphics[trim=2 2 2 2,clip, width=1\textwidth]{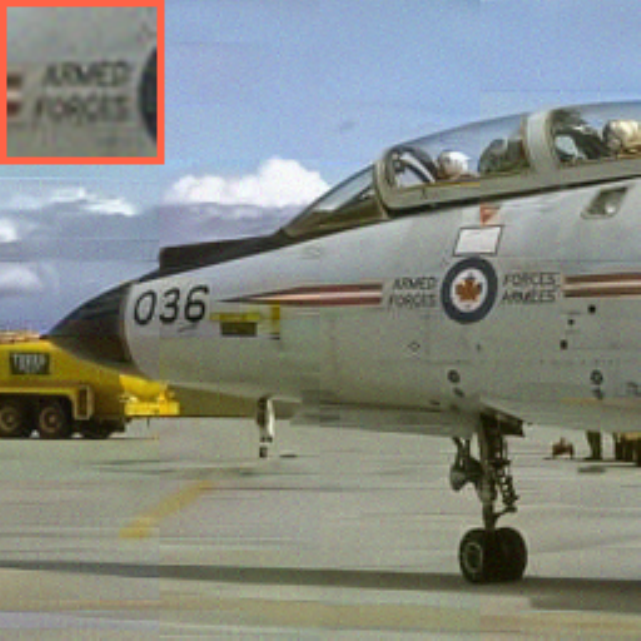}  
\centerline{ 31.04 dB, 0.912}
\end{minipage}\\

\begin{minipage}{.20\textwidth}
\vspace{0.04cm}
\includegraphics[trim=2 2 2 2,clip, width=1\textwidth]{spc/color/monarch/mr_05/orig.pdf}  
\centerline{}
\end{minipage}\hspace{0.1cm}
\begin{minipage}{.20\textwidth}
\vspace{0.04cm}
\includegraphics[trim=2 2 2 2,clip, width=1\textwidth]{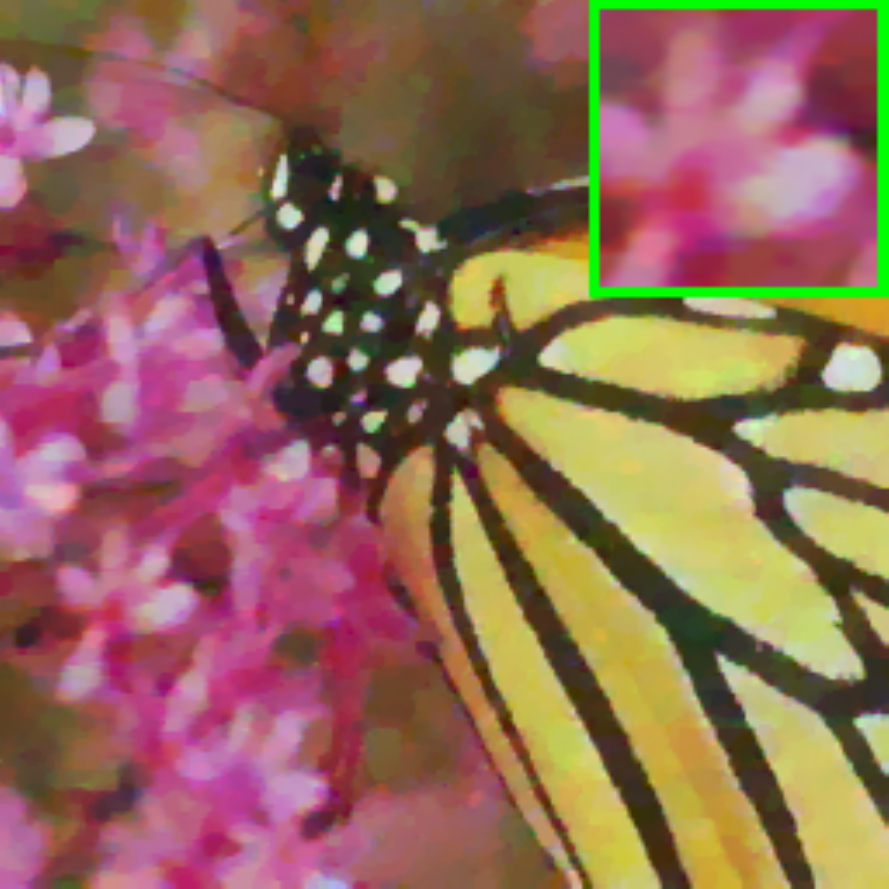}  
\centerline{ 23.78 dB, 0.917}
\end{minipage}\hspace{0.1cm}
\begin{minipage}{.20\textwidth}
\vspace{0.1cm}
\includegraphics[trim=2 2 2 2,clip, width=1\textwidth]{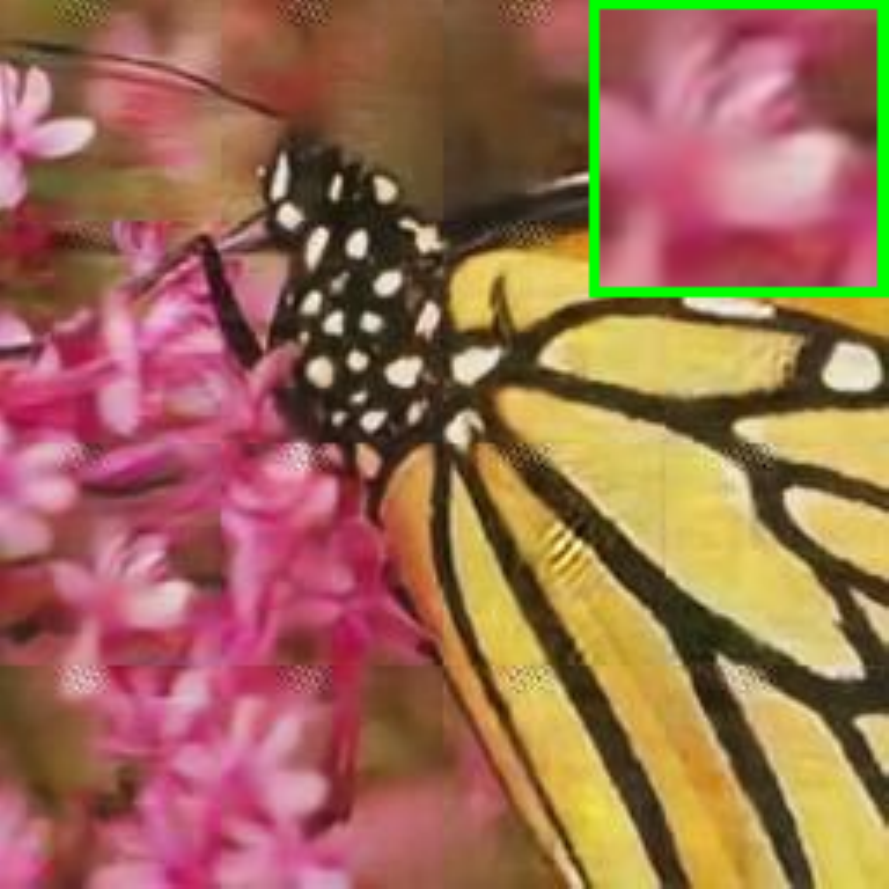}  
\centerline{ 25.72 dB, 0.952}
\end{minipage}\hspace{0.1cm}
\begin{minipage}{.20\textwidth}
\vspace{0.1cm}
\includegraphics[trim=2 2 2 2,clip, width=1\textwidth]{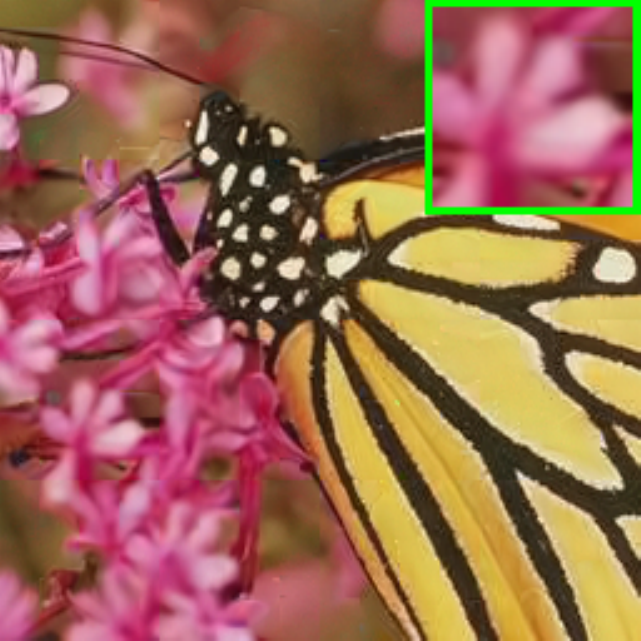}  
\centerline{ 32.14 dB, 0.987}
\end{minipage}
\end{minipage}

\centerline{}
\centerline{}

\begin{minipage}{.01\textwidth}
\raggedleft
\begin{turn}{90}Reconstructions with 40\% M.R. \end{turn}
\end{minipage}
\begin{minipage}{.98\textwidth}
\centering
\begin{minipage}{.20\textwidth}
\vspace{0.01cm}
\includegraphics[trim=2 2 2 2,clip, width=1\textwidth]{spc/color/plane/mr_05/orig.pdf}  
\centerline{}
\end{minipage}\hspace{0.1cm}
\begin{minipage}{.20\textwidth}
\vspace{0.01cm}
\includegraphics[trim=2 2 2 2,clip, width=1\textwidth]{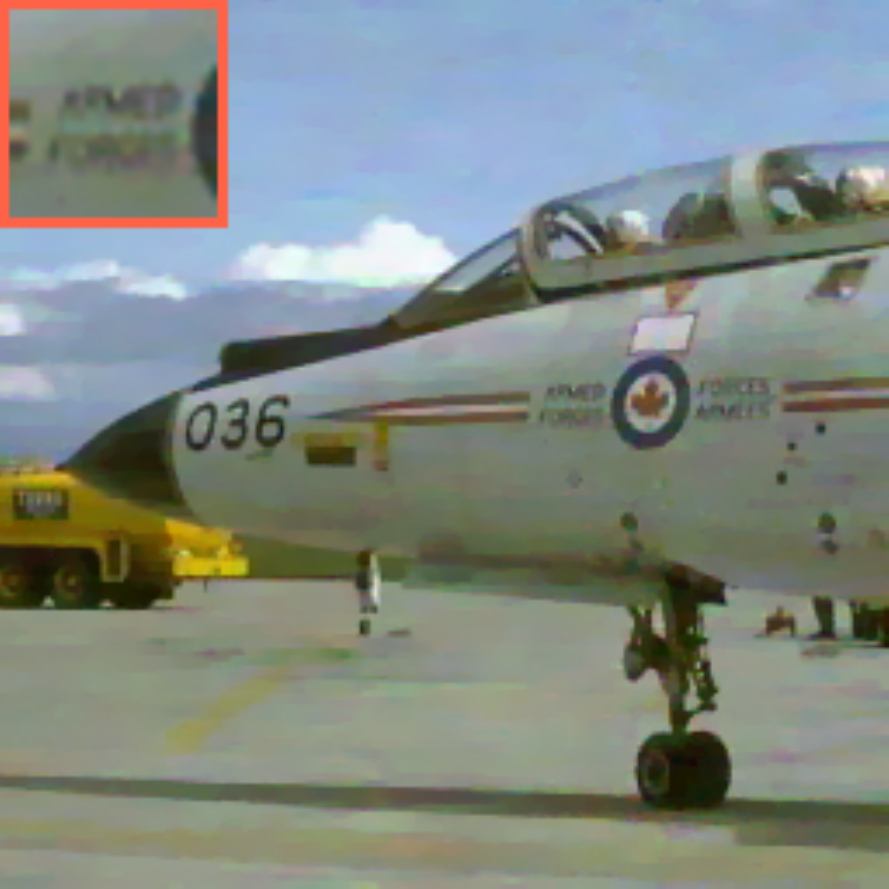}  
\centerline{ 29.86 dB, 0.911}
\end{minipage}\hspace{0.1cm}
\begin{minipage}{.20\textwidth}
\vspace{0.1cm}
\includegraphics[trim=2 2 2 2,clip, width=1\textwidth]{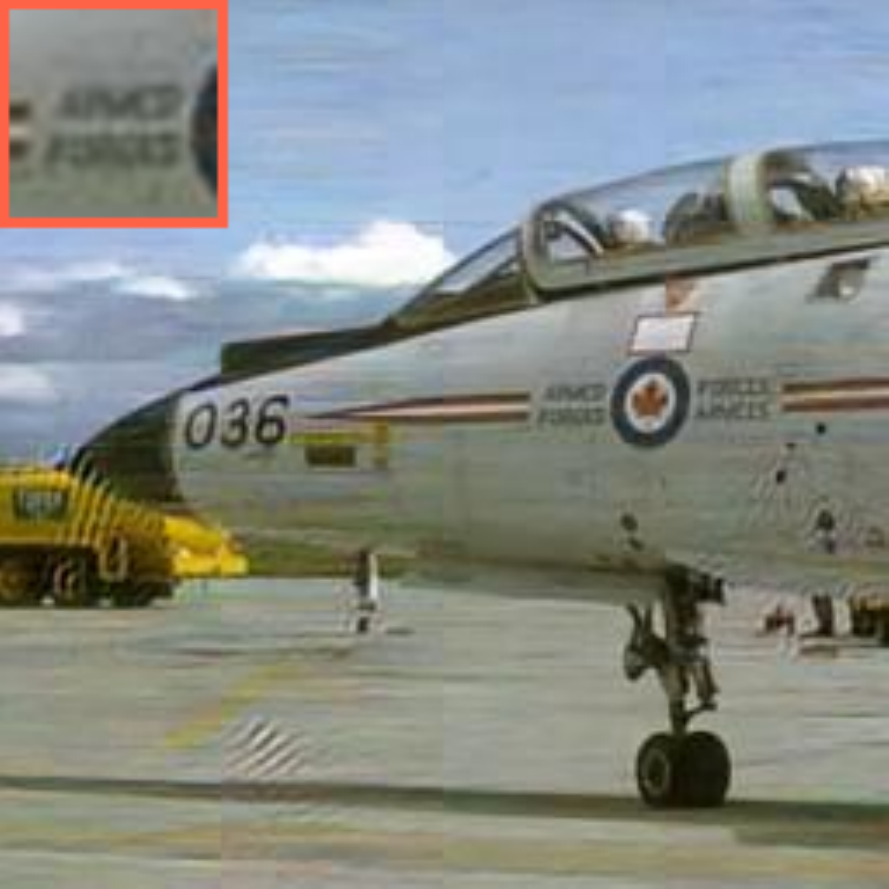}  
\centerline{ 29.22 dB, 0.910}
\end{minipage}\hspace{0.1cm}
\begin{minipage}{.20\textwidth}
\vspace{0.1cm}
\includegraphics[trim=2 2 2 2,clip, width=1\textwidth]{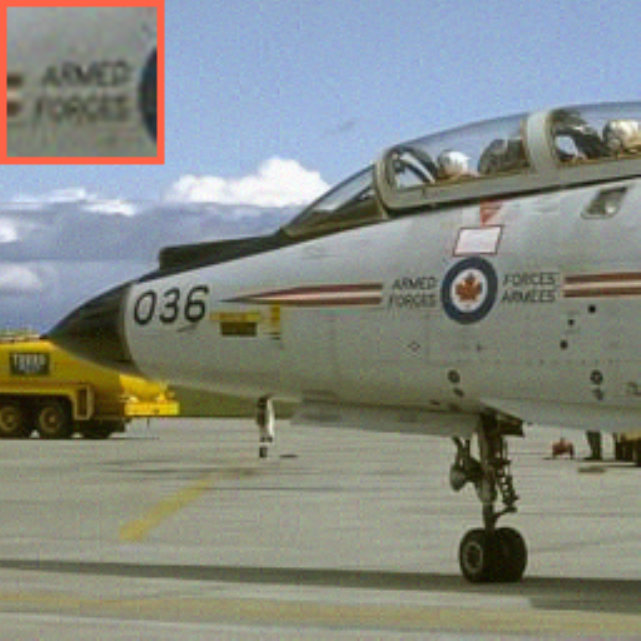}  
\centerline{ 33.10 dB, 0.936}
\end{minipage}

\begin{minipage}{.20\textwidth}
\vspace{-0.35cm}
\includegraphics[trim=2 2 2 2,clip, width=1\textwidth]{spc/color/monarch/mr_05/orig.pdf}  
\end{minipage}\hspace{0.1cm}
\begin{minipage}{.20\textwidth}
\vspace{0.1cm}
\includegraphics[trim=2 2 2 2,clip, width=1\textwidth]{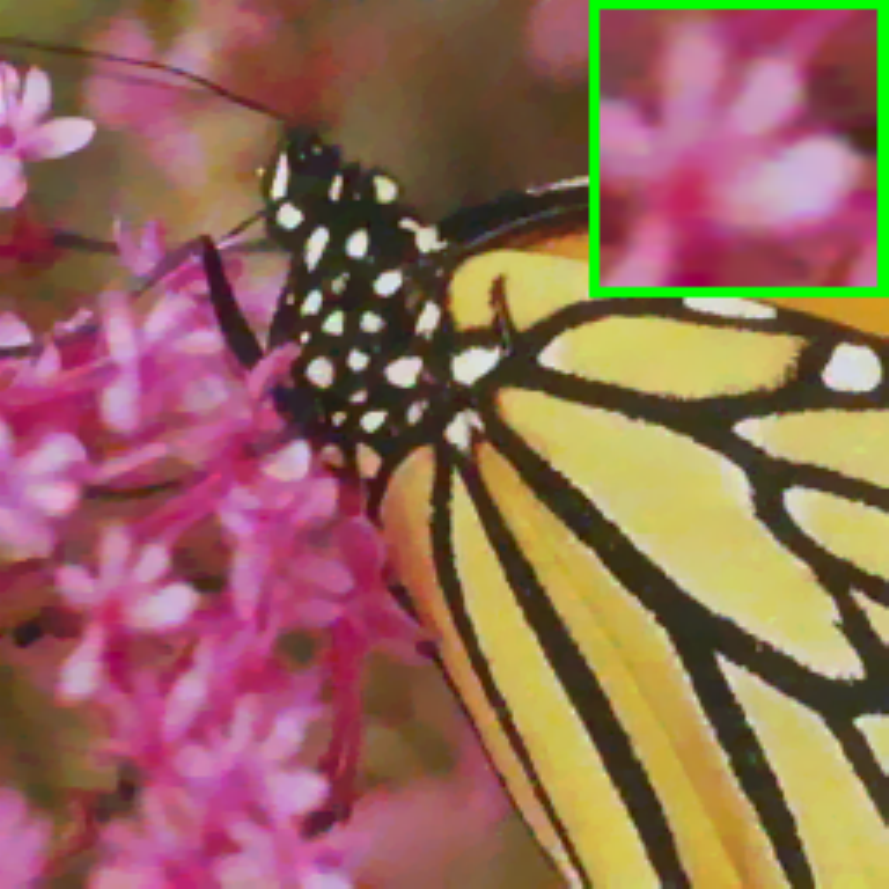}  
\centerline{ 27.07 dB, 0.957}
\end{minipage}\hspace{0.1cm}
\begin{minipage}{.20\textwidth}
\vspace{0.1cm}
\includegraphics[trim=2 2 2 2,clip, width=1\textwidth]{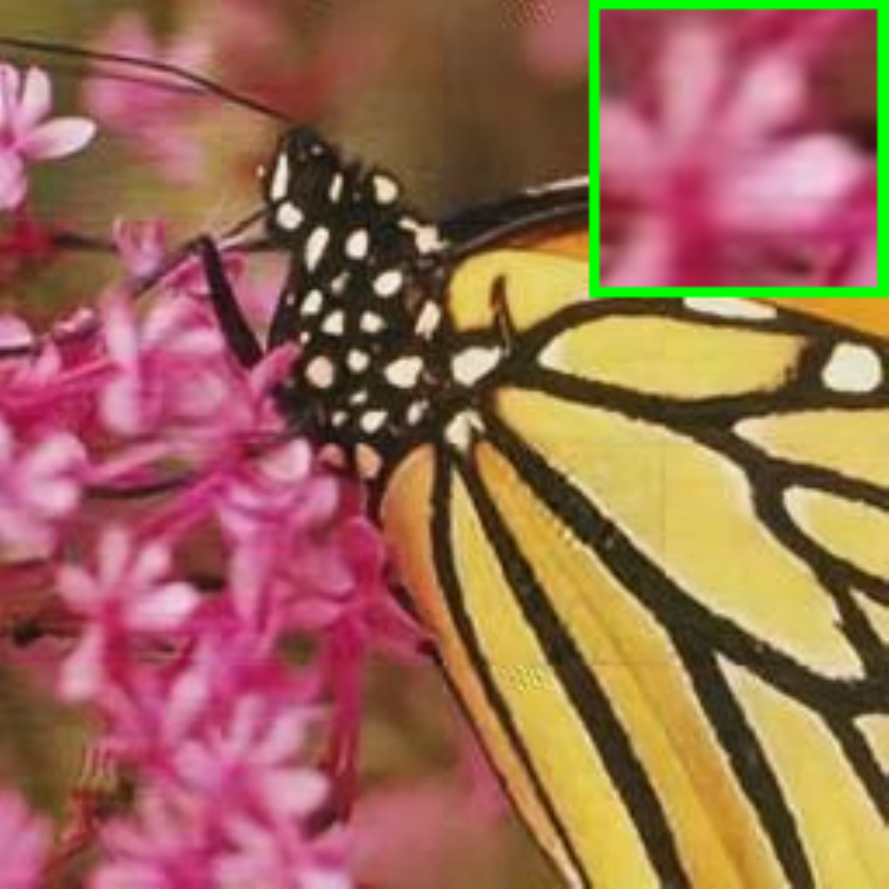}  
\centerline{ 29.94 dB, 0.979}
\end{minipage}\hspace{0.1cm}
\begin{minipage}{.20\textwidth}
\vspace{0.1cm}
\includegraphics[trim=2 2 2 2,clip, width=1\textwidth]{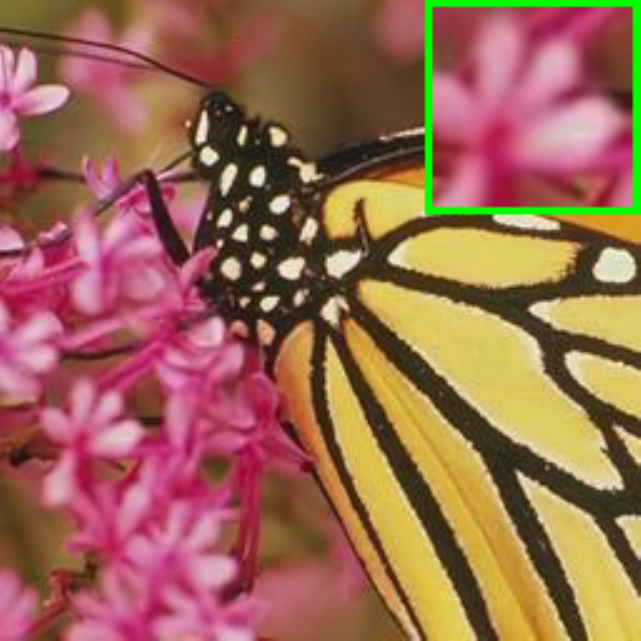}  
\centerline{ 37.81 dB, 0.996}
\end{minipage}
\end{minipage}
\caption{ Qualitative comparisons of images reconstructed from simulated LiSens measurements using TVAL3, OneNet and our approach. Reconstructions from our approach have minimal artifacts and are closer to the original image. }
\label{fig:lisens_color}
\end{figure*}

 The reconstruction in case of simulated  LiSens is done at 25\% and 40\% measurement rates. Our LiSens experiments, similar to SPC experiments, have been done on both 128x128 and 256x256 images as shown in Table \ref{table:licens_sim} and Figure \ref{fig:lisens_color} respectively. We compare our reconstructions with that obtained using OneNet. Our method provides better results in terms of visual perception as well as PSNR and SSIM values. Our reconstructions have well-defined boundaries of different objects in the image and do not produce artifacts which are observed in case of OneNet. We have used hard constraint case for the simulated LiSens reconstruction for approximately 2000 iterations with a step-size of 7.5.

\begin{table}[!h]
\renewcommand{\arraystretch}{1.3}
\caption{Comparisons of reconstructions from simulated LiSens measurements at different measurement rates for the images shown in Figure \ref{fig:sel_imgs}.Our approach obtains better performance than OneNet and TVAL3. See Figure \ref{fig:lisens_color} for qualitative comparisons}
\label{table:licens_sim}
\begin{center}
\begin{tabular}{cccccccc}

\toprule
    \multirow{2}{*}{ Name}&\multirow{2}{*}{M.R.}&\multicolumn{2}{c}{TVAL3}&\multicolumn{2}{c}{OneNet}&\multicolumn{2}{c}{Ours}\\
    \cline{3-8} & &  PSNR & SSIM & PSNR & SSIM  & PSNR & SSIM  \\
\midrule 
 \multirow{2}{*}{bird} & 25 & 24.59 & 0.95 & 24.98 & 0.82 & \textbf{27.13} & \textbf{0.96} \\
& 40 & 29.34 & 0.98 & 27.52 & 0.96 & \textbf{34.14} & \textbf{0.99} \\
\midrule 
 \multirow{2}{*}{building} & 25 & 18.72 & 0.67 & 21.16 & 0.79 & \textbf{30.87} & \textbf{0.95} \\
& 40 & 23.41 & 0.82 & 22.41 & 0.84 & \textbf{35.06} & \textbf{0.98} \\
\midrule 
 \multirow{2}{*}{cat} &25 & 23.41 & 0.67 & 27.27 & 0.89 & \textbf{29.95} & \textbf{0.94} \\
& 40 & 25.83 & 0.87 & 29.03 & 0.92 & \textbf{34.65} & \textbf{0.97} \\
\midrule 
 \multirow{2}{*}{flower} &25 & 21.00 & 0.72 & \textbf{27.85} & \textbf{0.91} & 26.54 & 0.88 \\
& 40 & 23.66 & 0.83 & \textbf{30.79} & \textbf{0.95} & 30.21 & 0.93 \\
\midrule 
 \multirow{2}{*}{parrot} &25 & 15.27 & 0.65 & 26.02 & 0.90 & \textbf{30.17} & \textbf{0.94} \\
& 40 & 19.75 & 0.85 & 27.99 & 0.93 & \textbf{32.35} & \textbf{0.96} \\
\midrule 
 \multirow{2}{*}{mean} &25 & 20.60 & 0.73 & 25.45  & 0.89  & \textbf{28.93} &  \textbf{0.94}\\
& 40 & 24.40 & 0.87 & 27.55  & 0.92 & \textbf{33.28} &  \textbf{0.97} \\
\bottomrule
\end{tabular}
\end{center}
\end{table}

\subsubsection{Real measurements}
 The real LiSens experiments have been done at 16\% and 33\% measurement rates obtained at a resolution of $768\times256$ , as provided by the authors of \cite{wang2015lisens}. We compare our real Lisens with TVAL3 as in Figure \ref{fig:lisens_real}. Our method performs better reconstruction with respect to low level details in the image. Our proposed method’s reconstruction has little or no blur compared to TVAL3 and the reconstruction is sharper in terms of object boundaries in the image. We use Hard constraint method for reconstruction with 25\% dropout in pixel-wise update. We use an update step of 7.5 and 2000 iterations for reconstruction, similar to simulated experiment. 

\subsection{FlatCam}

\begin{figure*}[!t]
    \centering
\begin{minipage}{\textwidth}
\centering
\begin{minipage}{.18\textwidth}
\centerline{Original image}
\vspace{0.01cm}
\includegraphics[trim=2 2 2 2,clip, width=1\textwidth]{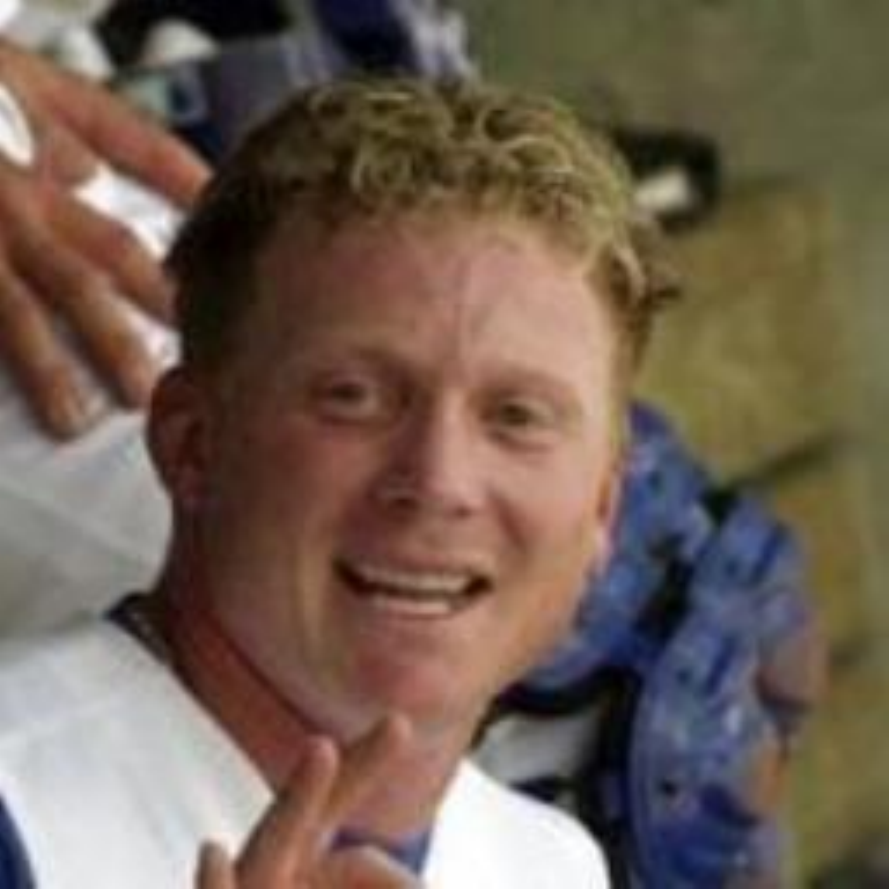}  
\centerline{}
\end{minipage}\hspace{0.1cm}
\begin{minipage}{.18\textwidth}
\centerline{L2 Reg.}
\vspace{0.1cm}
\includegraphics[trim=2 2 2 2,clip, width=1\textwidth]{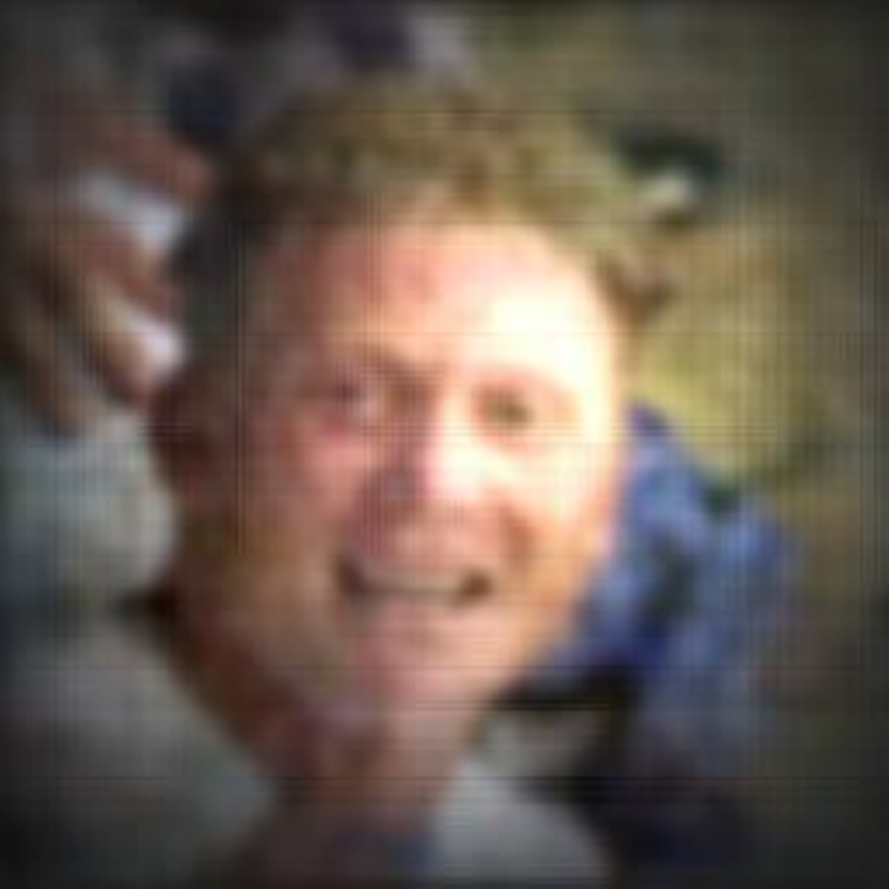}  
  
\centerline{ 12.79 dB, 0.71}
\end{minipage}\hspace{0.1cm}
\begin{minipage}{.18\textwidth}
\centerline{OneNet}
\vspace{0.1cm}
\includegraphics[trim=2 2 2 2,clip, width=1\textwidth]{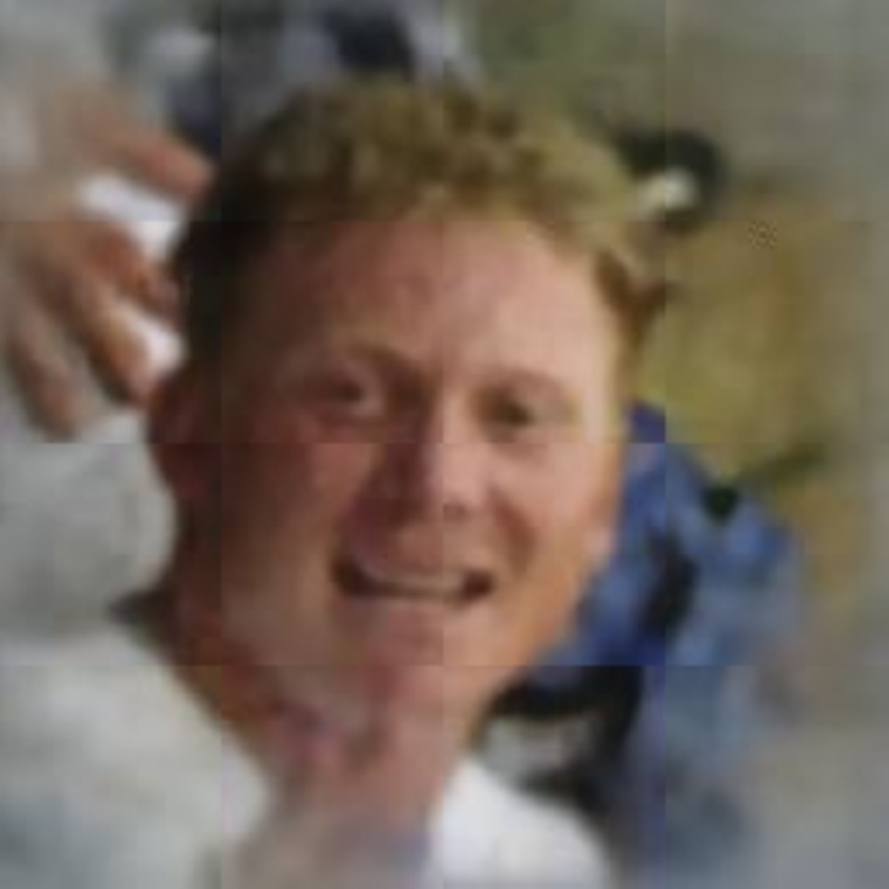}  
  
\centerline{ 20.11 dB, 0.84}
\end{minipage}\hspace{0.1cm}
\begin{minipage}{.18\textwidth}
\centerline{Ours}
\vspace{0.1cm}
\includegraphics[trim=2 2 2 2,clip, width=1\textwidth]{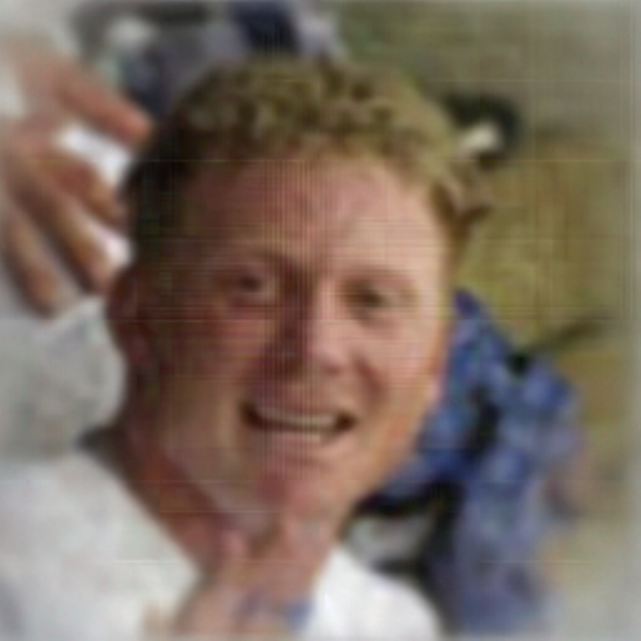}  
\centerline{ 20.22 dB, 0.85}

\end{minipage}\\

\end{minipage}

\begin{minipage}{\textwidth}
\centering
\begin{minipage}{.18\textwidth}

\vspace{0.01cm}
\includegraphics[trim=2 2 2 2,clip, width=1\textwidth]{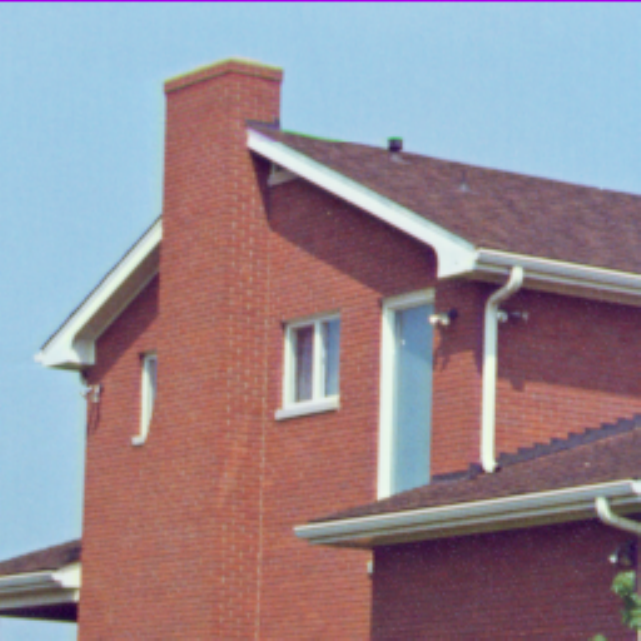}  
\centerline{}
\end{minipage}\hspace{0.1cm}
\begin{minipage}{.18\textwidth}
\vspace{0.1cm}
\includegraphics[trim=2 2 2 2,clip, width=1\textwidth]{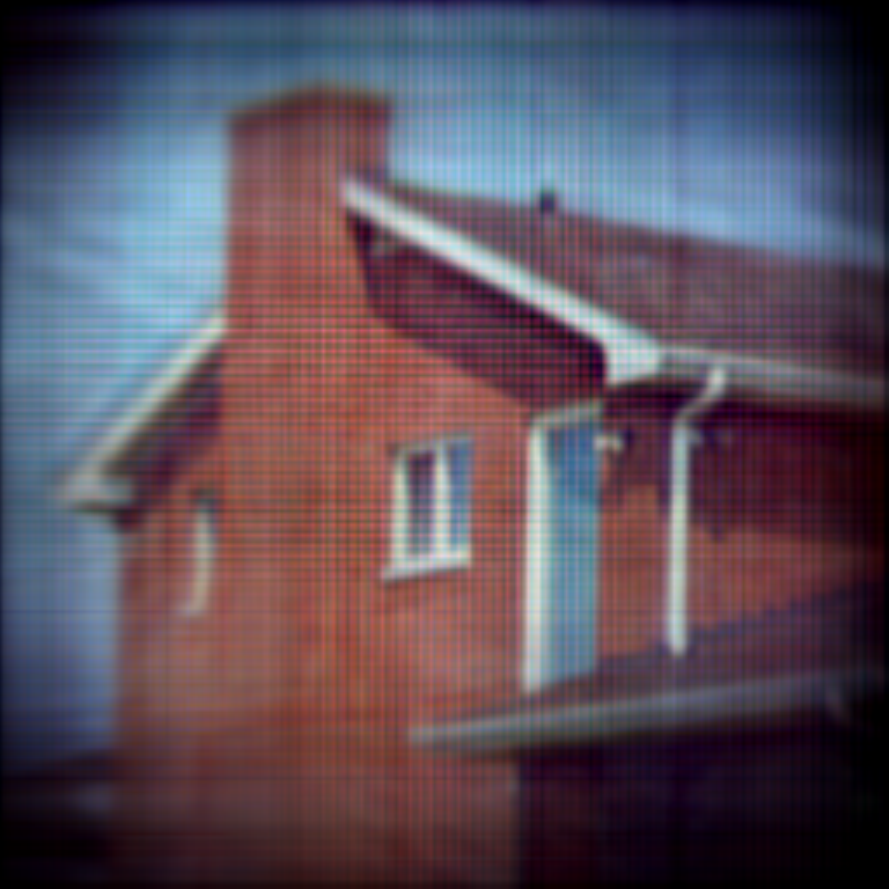}  
  
\centerline{ 10.77 dB, 0.59}
\end{minipage}\hspace{0.1cm}
\begin{minipage}{.18\textwidth}
\vspace{0.1cm}
\includegraphics[trim=2 2 2 2,clip, width=1\textwidth]{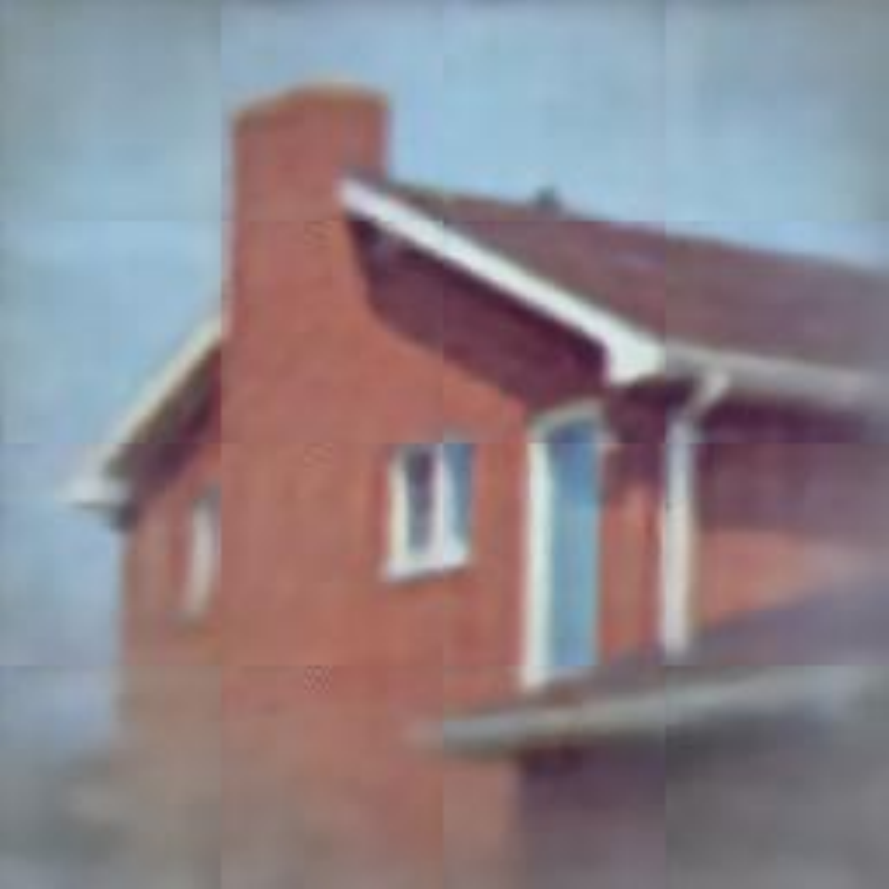}  
  
\centerline{ 19.52 dB, 0.81}
\end{minipage}\hspace{0.1cm}
\begin{minipage}{.18\textwidth}
\vspace{0.1cm}
\includegraphics[trim=2 2 2 2,clip, width=1\textwidth]{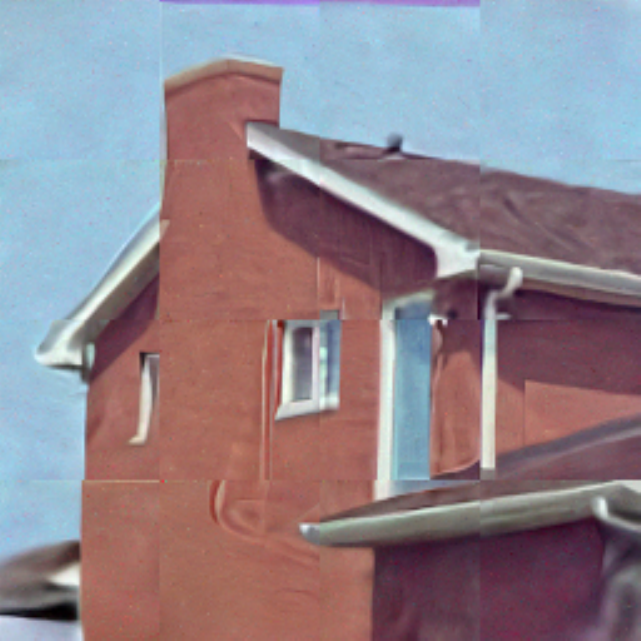}  
\centerline{ 25.08 dB, 0.83}
\end{minipage}\\

\end{minipage}
    \caption{ Reconstructions ($256\times 256$) from simulated FlatCam measurements using L2 regularization, OneNet and our approach. Note the suppression of vignetting effect in our results which is clearly visible in the house image.}
    \label{fig:flatcam_color}
\end{figure*}

\subsubsection{Simulation case}
The matrices $\Phi_L$ and $\Phi_R$ in the FlatCam imaging model are estimated based on the calibration procedure mentioned in \cite{asif2017flatcam}. As we want to deal with RGB images, separate $\Phi_L$ and $\Phi_R$ matrices are calibrated for each of the R, G and B channels with the help of a Bayer color filter array on the sensor. We compare our results with OneNet and L2 regularisation, on two 256x256 images as shown in Figure \ref{fig:flatcam_color}. Our method shows better PSNR, SSIM, and perceptually better quality samples. Our method produces the least blurry solution and objects in the image has well defined boundaries. We use 25$\%$ pixel dropout and perform 1000 iterations of augmented Lagrangian method with the step size $\alpha$ as $60.0$ and $\rho$ as 10. 

\begin{figure}[!t]
    \centering
    \begin{minipage}{0.5\textwidth}
    \begin{minipage}{0.32\textwidth}
    \includegraphics[width=\textwidth]{flatcam/color/lfw/orig.pdf}
    \end{minipage}
    \begin{minipage}{0.32\textwidth}
    \includegraphics[width=\textwidth]{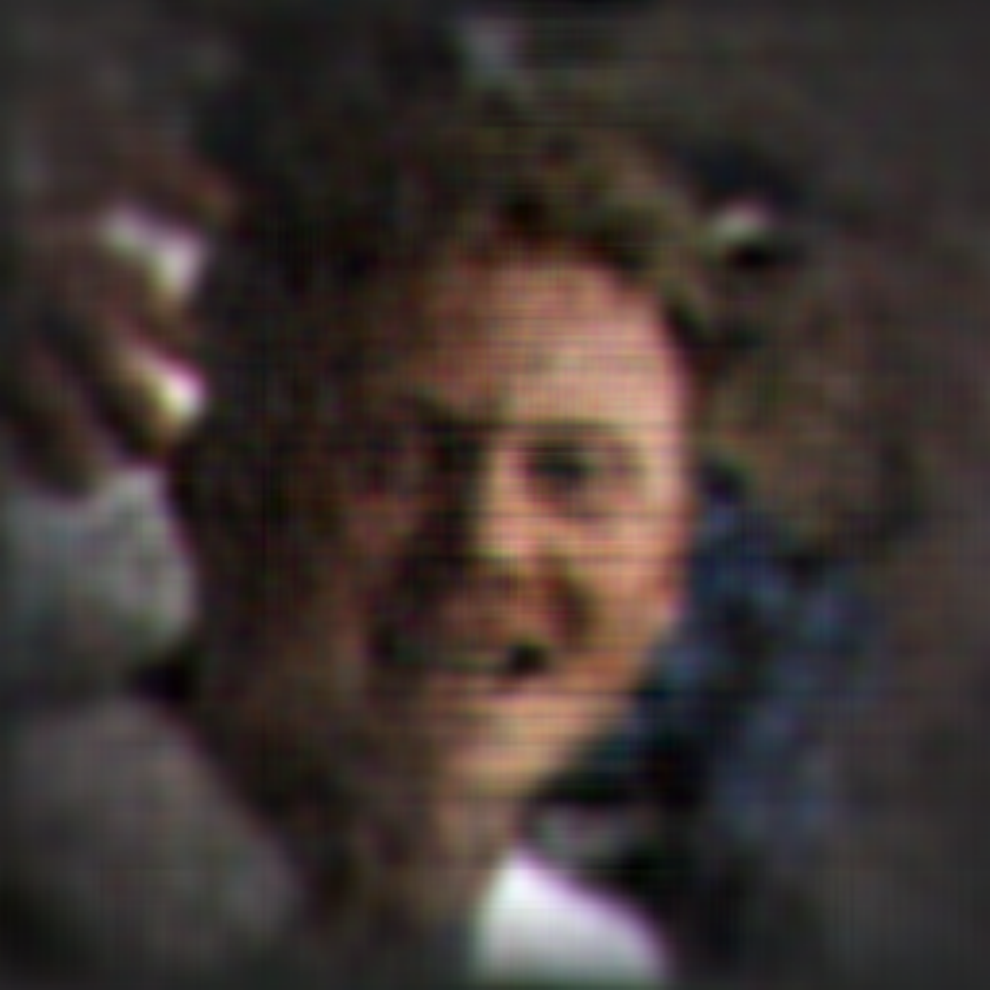}
    \end{minipage}
    \begin{minipage}{0.32\textwidth}
    \includegraphics[width=\textwidth]{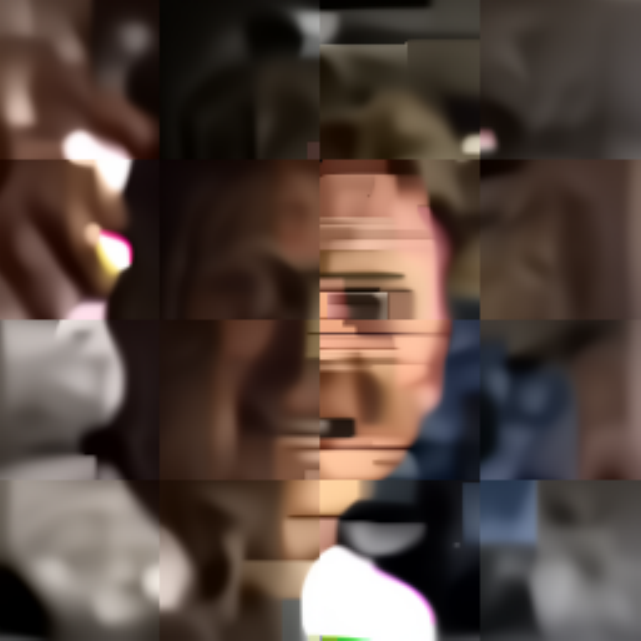}
    \end{minipage}\\\\
    
    \begin{minipage}{0.32\textwidth}
    \includegraphics[width=\textwidth]{flatcam/color/house/orig.pdf}
    \centerline{Original Image}
    \end{minipage}
    \begin{minipage}{0.32\textwidth}
    \includegraphics[width=\textwidth]{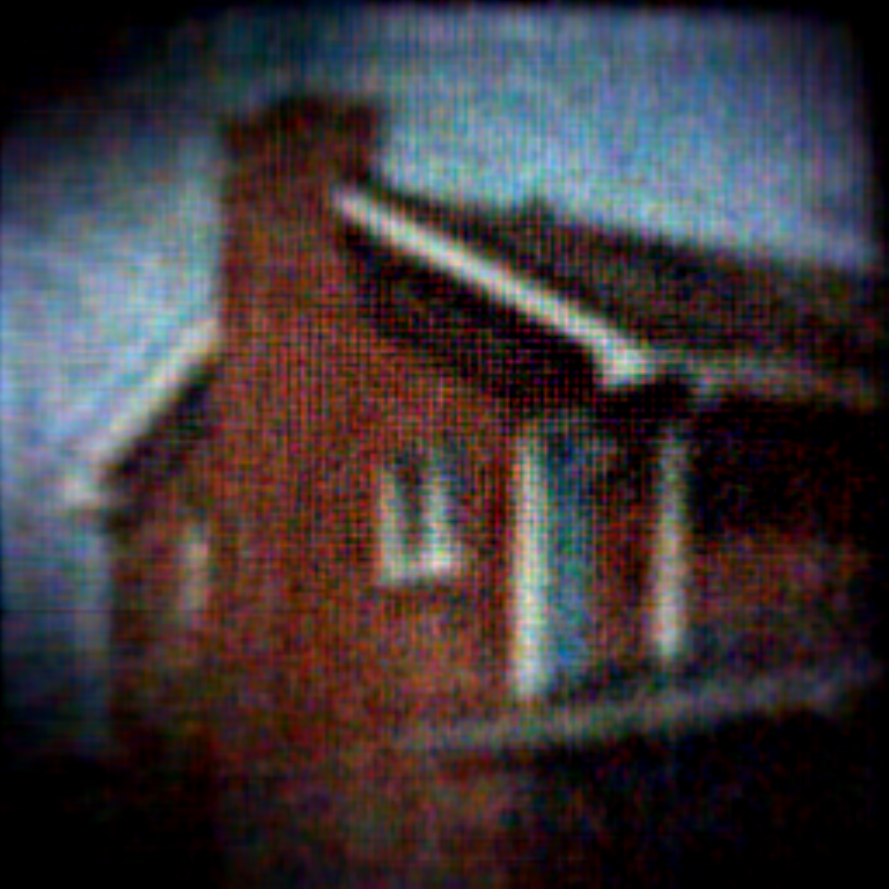}
    \centerline{L2 regularization}
    \end{minipage}
    \begin{minipage}{0.32\textwidth}
    \includegraphics[width=\textwidth]{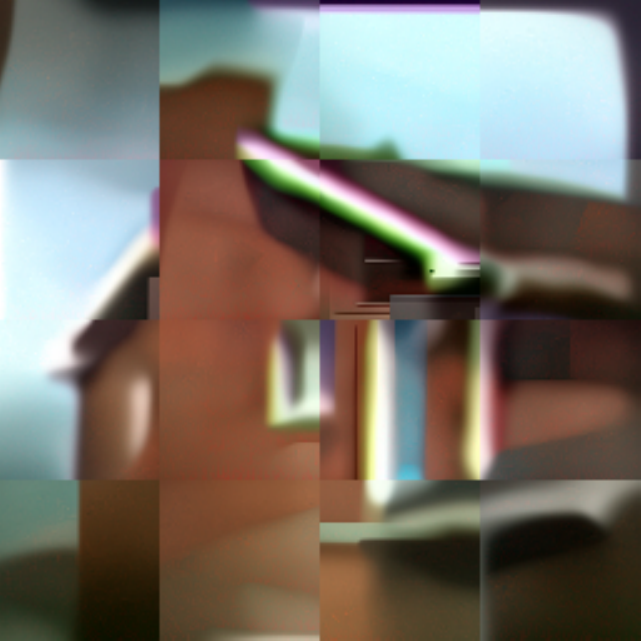}
    \centerline{Ours}
    \end{minipage}
    \end{minipage}
    \caption{Qualitative comparisons of reconstructions obtained from real FlatCam measurements using calibrated $\Phi_L$ and $\Phi_R$ using L2 regularization and our approach. Real reconstructions are not good because of calibration error and separability assumption in the forward model.}
    \label{fig:flatcam_real}
\end{figure}

\subsubsection{Real measurements}
We use the data provided by the authors of \cite{asif2017flatcam}. The original images were displayed on a monitor and captured using FlatCam. Using a Bayer color filter on the sensor, separate measurements for the three color channels can be obtained. We compare our reconstructions with L2 regularization as shown in Figure \ref{fig:flatcam_real}. Our reconstructions are more accurate in terms of brightness, boundaries and sharpness of the image. We use soft constraint case for reconstruction and use the same hyperparameters as in the simulation case. 

\begin{figure}[!t]
\begin{minipage}{.50\textwidth}
\centering
\centerline{Reconstructions with 16\% M.R.}
\vspace{0.1 cm}
\begin{minipage}{.30\textwidth}
\includegraphics[trim=0 300 0 0,clip, width=1\textwidth]{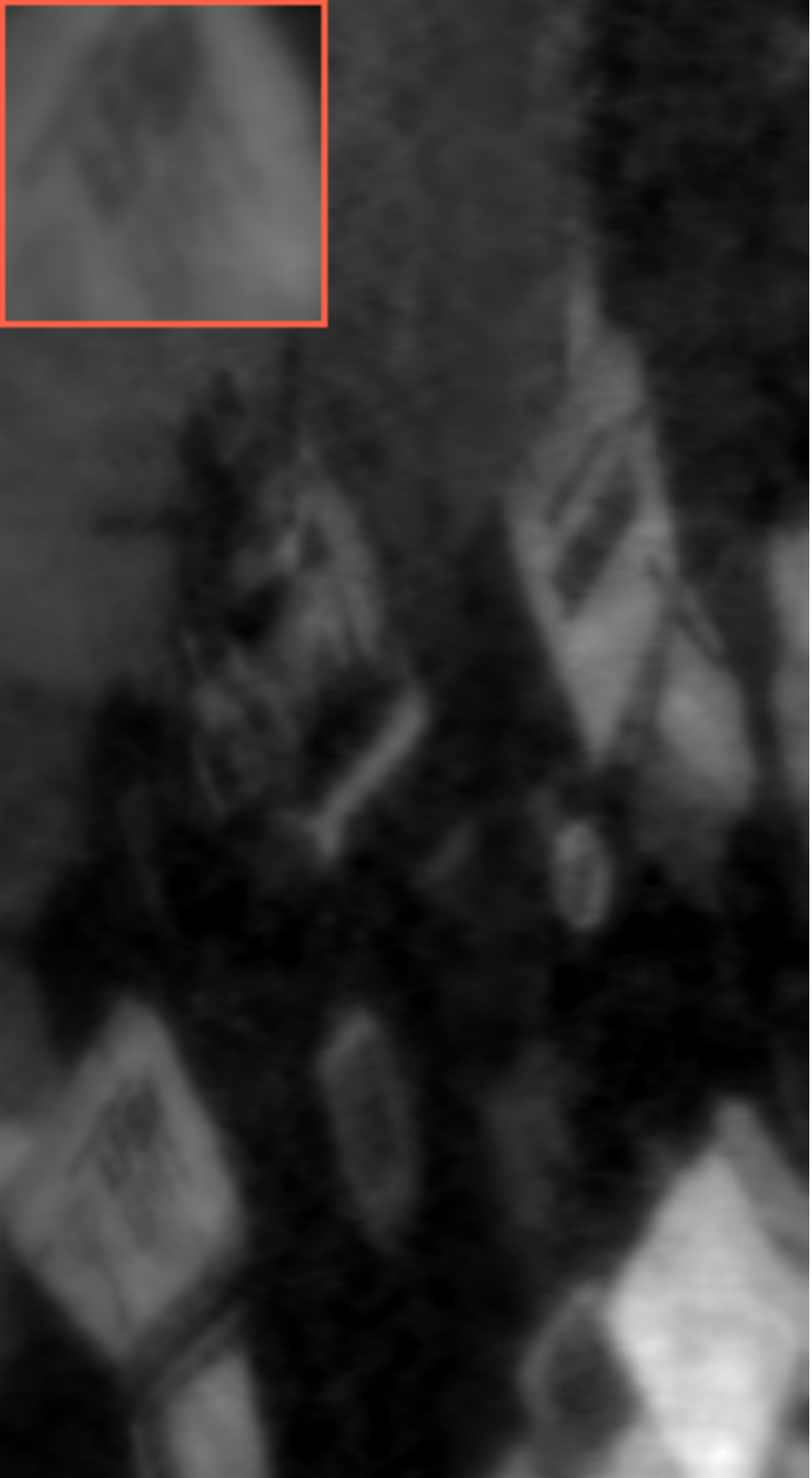}  
\end{minipage}
\begin{minipage}{.30\textwidth}
\includegraphics[trim=0 300 0 0,clip, width=1\textwidth]{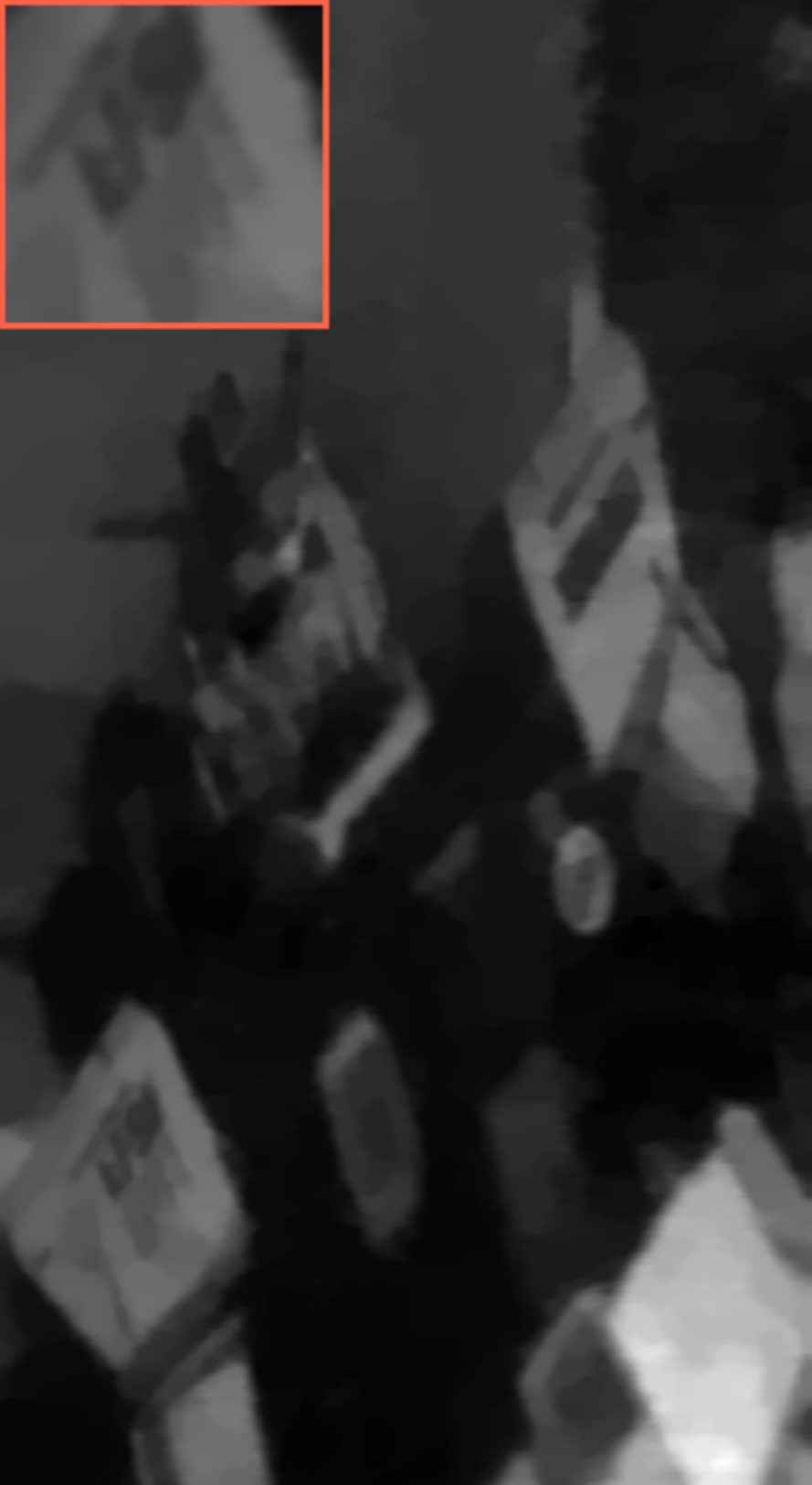}  
\end{minipage}
\begin{minipage}{.30\textwidth}
\includegraphics[trim=0 120 0 0,clip, width=1\textwidth]{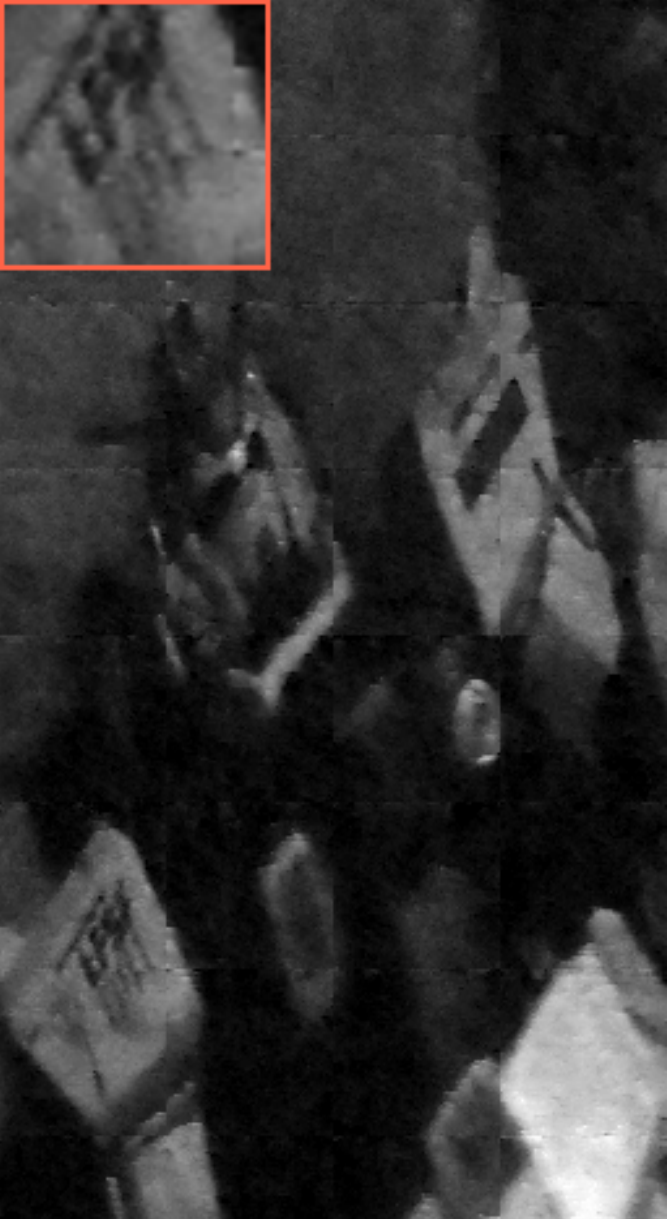}  
\end{minipage}\\
\vspace{0.1cm}
\centerline{Reconstructions with 33\% M.R.}
\vspace{0.1cm}
\begin{minipage}{.30\textwidth}
\includegraphics[trim=0 300 0 0,clip, width=1\textwidth]{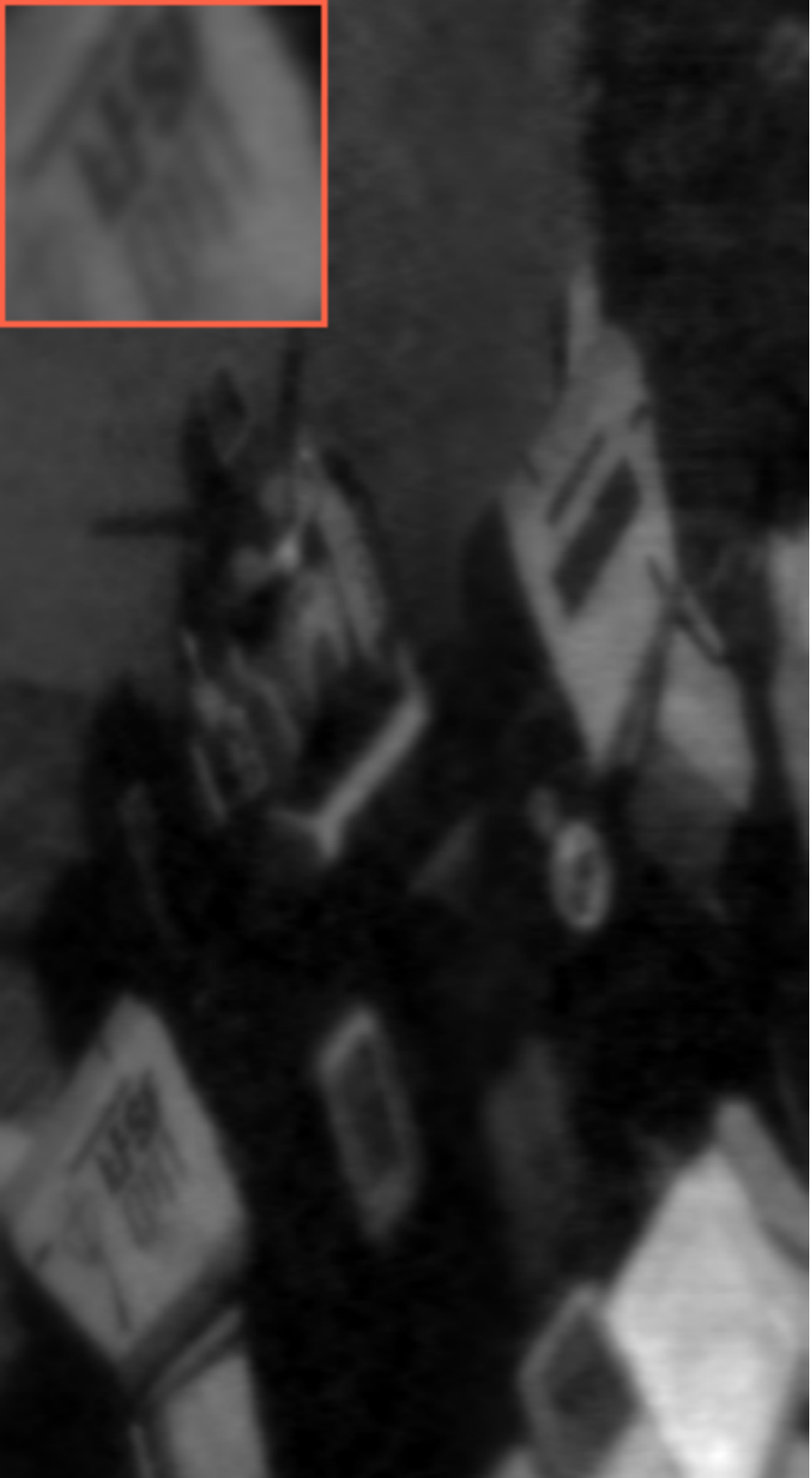}  
\centerline{Gradient L2 reg.}
\end{minipage}
\begin{minipage}{.30\textwidth}
\includegraphics[trim=0 300 0 0,clip, width=1\textwidth]{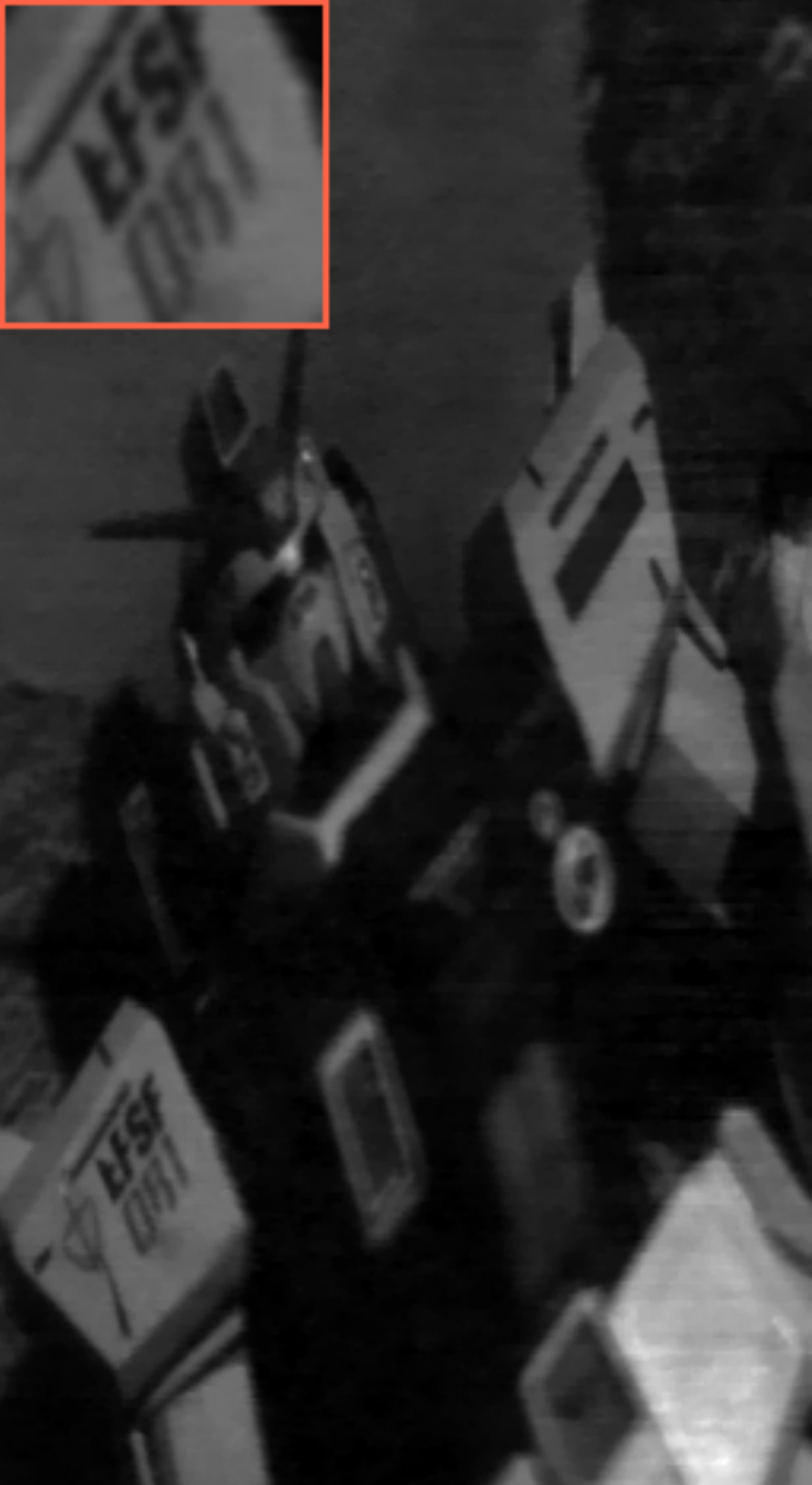}
\centerline{TVAL3}
\end{minipage}
\begin{minipage}{.30\textwidth}
\includegraphics[trim=0 120 0 0,clip, width=1\textwidth]{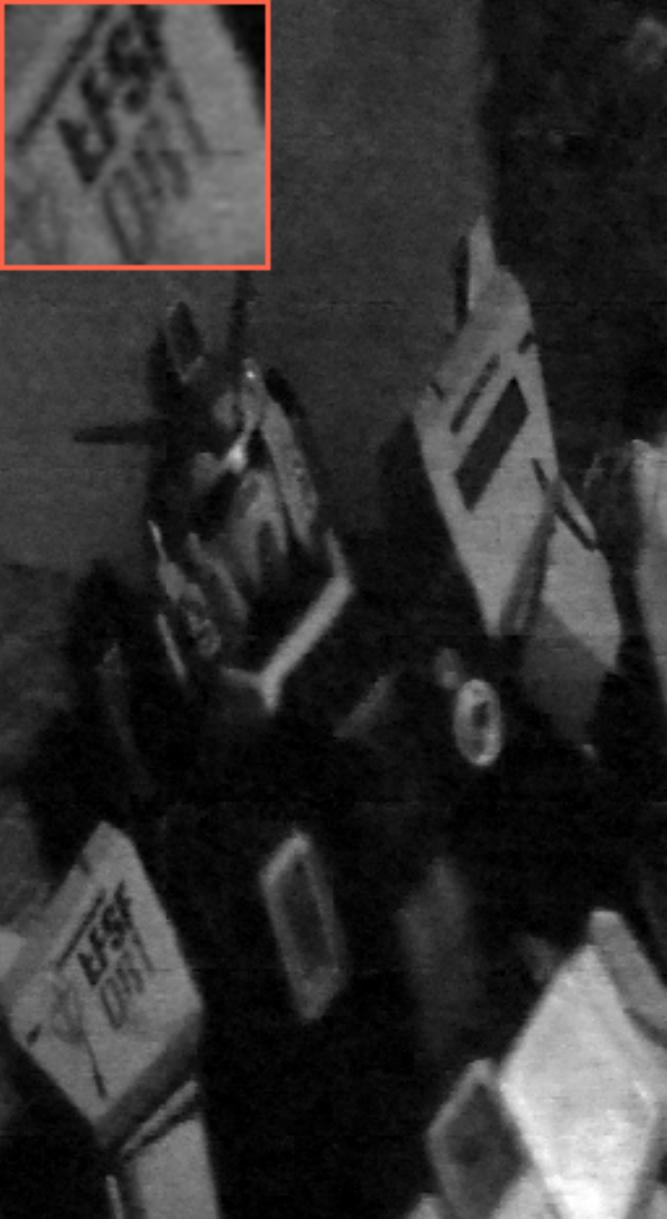} 
\centerline{Ours}
\end{minipage}
\end{minipage}
\caption{Qualitative comparisons of reconstructions from real LiSens measurements at different measurement rates. Reconstructions using our method are sharper and preserve the overall structure. }
\label{fig:lisens_real}
\end{figure}
We observe that reconstructions from real FlatCam are not qualitatively as good as with real SPC and LiSens measurements. This is because the forward model assumed in this case is erroneous. Firstly, there are calibration errors in estimating the $\Phi_L$ and $\Phi_R$ matrices. Secondly, the forward model in \cite{asif2017flatcam} relies on the separability assumption leading to model error. 

\subsection{Ablation Experiments}

\subsubsection{Effect of pixel-wise dropout}
\label{sec:abalation}
\begin{figure}[!h]
    \centering
\begin{minipage}{0.5\textwidth}
\centering
\begin{minipage}{.30\textwidth}
\centerline{Original image}
\vspace{0.01cm}
\includegraphics[trim=2 2 2 2,clip, width=1\textwidth]{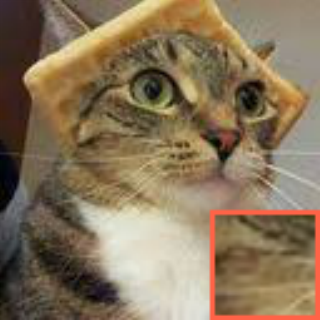}  
\centerline{}
\end{minipage}\hspace{0.1cm}
\begin{minipage}{.30\textwidth}
\centerline{0\% pixel dropout}
\vspace{0.1cm}
\includegraphics[trim=2 2 2 2,clip, width=1\textwidth]{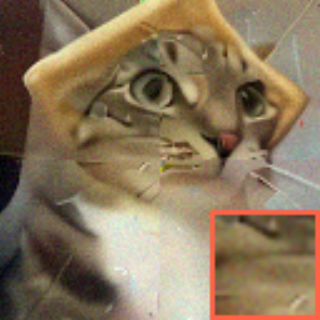}  
  
\centerline{ 26.34 dB, 0.826}
\end{minipage}\hspace{0.1cm}
\begin{minipage}{.30\textwidth}
\centerline{25\% pixel dropout}
\vspace{0.1cm}
\includegraphics[trim=2 2 2 2,clip, width=1\textwidth]{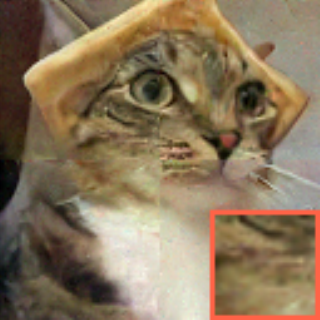}

\centerline{ 27.93 dB, 0.887}
\end{minipage}\hspace{0.1cm}\\
\begin{minipage}{.30\textwidth}
\centerline{50\% pixel dropout}
\vspace{0.1cm}
\includegraphics[trim=2 2 2 2,clip, width=1\textwidth]{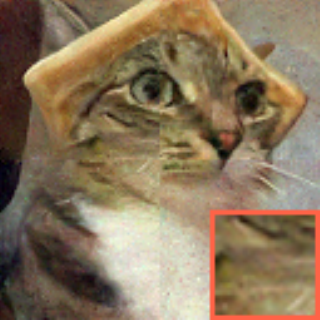}  
  
\centerline{ 26.82 dB, 0.856}
\end{minipage}\hspace{0.1cm}
\begin{minipage}{.30\textwidth}
\centerline{75\% pixel dropout}
\vspace{0.1cm}
\includegraphics[trim=2 2 2 2,clip, width=1\textwidth]{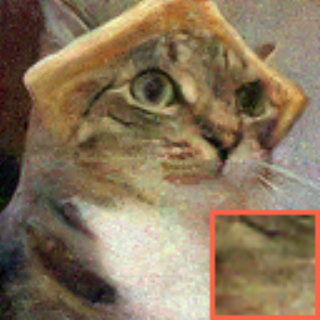}  
\centerline{ 25.87 dB, 0.820}

\end{minipage}\\

\end{minipage}
    \caption{Effect of varying the amount of pixel dropout for SPC reconstruction at 15\% measurement rate. By not updating a certain amount of pixels every iteration, the texture is reconstructed better and the image has a higher quality. However, on increasing this dropout ratio more than a certain level, the reconstructions become noisy and the quality reduces.}
    \label{fig:abalation_dropout}
\end{figure}

In this experiment, we vary the amount of pixels not updated in each iteration and observe its effect on the reconstructed image, see Figure \ref{fig:abalation_dropout}. When the dropout ratio is zero, the area in the image having texture is over smooth. With considerable dropout ratio ($25\%$), the texture is reconstructed better amounting to a higher PSNR and SSIM. However, on increasing it further, the reconstructions appear noisy with a reduction in quality. Thus, for all our experiments, we used $25\%$ dropout.   

\subsubsection{Comparison with Ride-CS - grayscale SPC}
\begin{figure}[!h]
    \centering
\begin{minipage}{0.5\textwidth}
\centering
\begin{minipage}{.30\textwidth}
\centerline{Original image}
\vspace{0.01cm}
\includegraphics[trim=2 2 2 2,clip, width=1\textwidth]{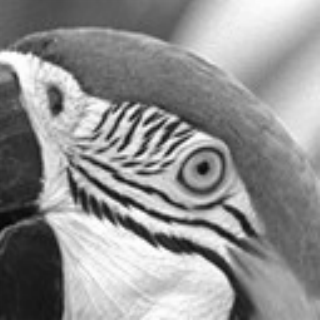}  
\centerline{}
\end{minipage}\hspace{0.1cm}
\begin{minipage}{.30\textwidth}
\centerline{RIDE-CS}
\vspace{0.1cm}
\includegraphics[trim=2 2 2 2,clip, width=1\textwidth]{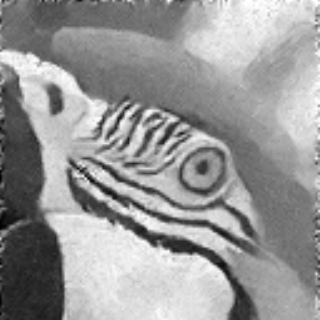}  
  
\centerline{ 23.01 dB, 0.697}
\end{minipage}\hspace{0.1cm}
\begin{minipage}{.30\textwidth}
\centerline{Ours}
\vspace{0.1cm}
\includegraphics[trim=2 2 2 2,clip, width=1\textwidth]{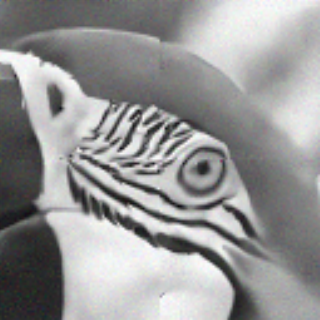}

\centerline{ 25.91 dB, 0.865}
\end{minipage}\hspace{0.1cm}
\end{minipage}

    \caption{Comparison with Ride-CS on reconstruction of grayscale image from simulated Single Pixel Camera measurements at 15\% measurement rate. Our reconstructions have a higher quality and are perceptually more closer to the true image.}
    \label{fig:abalation_ride}
\end{figure}

While we train our model on colored Imagenet data, we observe that in practice this approach works well on reconstructing grayscale images as well. We compare our reconstruction with that of RIDE-CS \cite{dave2017compressive}, which uses the autoregressive model RIDE \cite{theis2015generative} as image prior. In Figure \ref{fig:abalation_ride}, we compare the reconstruction of a grayscale image from Single Pixel Camera measurements using our approach and RIDE-CS for $15\%$ measurement rate. The reconstruction obtained from our approach is better than that of RIDE-CS. This is because we use PixelCNN++ which is a deeper network than RIDE and hence has better representation power.  Also, the running time of our approach ($\sim$ $5$ minutes ) is much less than that of RIDE-CS ($\sim$ $30$ minutes). Our approach is CNN based and hence can be parallelized over multiple GPUs while RIDE-CS relies on a network of spatial LSTMs which are tough to parallelize. 

\subsubsection{Comparison with OneNet in their original setting}
Till now we have performed all the experiments with different $\Phi$ matrix for each color channel. However in OneNet \cite{chang2017one}, the authors have considered one $\Phi$ matrix that multiplexes across the three color channels, which might not be feasible to implement in a real system. For this ablation experiment, we consider the original setting as used in \cite{chang2017one} and compare their reconstructions with ours for 10\% SPC reconstruction on the 9 test ImageNet images mentioned in the \cite{chang2017one}. PSNR and SSIM values for the same are mentioned in Table \ref{table:abalation_onenet}. Our approach performs better than OneNet. 
\begin{table}[!h]
\renewcommand{\arraystretch}{1.3}
\caption{Comparisons of compressive imaging reconstructions for images provided in \cite{chang2017one} with their setting of multiplexing across color channels. However, this way of multiplexing across the color channels might not be feasible in a real system.}
\label{table:abalation_onenet}
\begin{center}

\begin{tabular}{ccccc}

\toprule
    \multirow{2}{*}{Figure Name}&\multicolumn{2}{c}{OneNet}&\multicolumn{2}{c}{Ours}\\
    \cline{2-5} &   PSNR & SSIM & PSNR & SSIM  \\
\midrule

ball &	24.696	& 0.9023	& \textbf{26.656}	& \textbf{0.9300} \\
dalmatian &	20.650	& 0.8314	& \textbf{21.812}	& \textbf{0.8518} \\
dog &	26.873	& 0.8734	& \textbf{28.552}	& \textbf{0.8952} \\
field &	26.470	& 0.9112	& \textbf{29.017}	& \textbf{0.9149} \\
man &	29.152	& 0.9460	& \textbf{31.787}	& \textbf{0.9540} \\
mountain &	25.484	& 0.8821 & \textbf{28.993}	& \textbf{0.8912} \\
table &	19.397	& \textbf{0.8083}	& \textbf{20.955}	& 0.6662 \\
woman &	25.512	& 0.8518	& \textbf{27.321}	& \textbf{0.8906} \\
wolf &	25.976	& 0.8839	& \textbf{28.355}	& \textbf{0.9061} \\

\bottomrule
\end{tabular}
\end{center}
\end{table}

\section{Discussion and Conclusion}
We demonstrate the efficacy of deep pixel level image prior for ill-posed reconstruction in different computational imaging problems. Among the three proposed approaches for inference, hard and soft constraint based and ALM based, overall, soft constraint-based method works well and can handle noisy measurements by appropriately varying the tuning parameter, $\lambda$. However, when there is no noise or less noise in the measurements, the hard constraint-based method performs as good as soft constraint case with an additional advantage of being parameter free and hence is preferable. In fact, for our real experiments on SPC (Figure \ref{fig:spc_real}) and Lisens (Figure \ref{fig:lisens_real}), we use hard constraint-based inference, which produces reasonable results. For cases such as Flatcam, non-invertibility of $\Phi \Phi^T$ prevents the use of hard-constraint based inference.

Our approach enjoys the versatility of image priors and rich feature representation of deep neural networks. Being pixel level, it explicitly accounts for pixel level correlations resulting in consistent texture and edges. We show our evaluations on both the simulation of forward models and data from real setups. In all cases, both quantitative and qualitative metrics suggest that our approach performs better than traditional methods and current state-of-the-art learning based methods. An interesting line of work would be to incorporate deviations from the forward model, due to calibration and model errors, in our approach to further improve the quality of reconstruction for FlatCam. 

\section*{Acknowledgment}
This work is supported by Qualcomm Innovation Fellowship (QInF) 2016 and 2017. We would like to thank Dr. Aswin Sankaranarayanan and Jian Wang from CMU  for sharing the real measurements for SPC and LiSens setup. We would like to thank Dr. Ashok Veeraraghavan, Vivek Boominathan, Jasper Tan from Rice University for sharing the FlatCam data and for useful discussions. 

\ifCLASSOPTIONcaptionsoff
  \newpage
\fi




%
\bibliographystyle{IEEEbib1}
\bibliography{refs}
\end{document}